\documentclass[sigconf]{acmart}

\usepackage{hyperref}
\usepackage{balance}
\usepackage{multirow}
\usepackage{booktabs}
\usepackage{url}
\usepackage{makecell}
\usepackage{colortbl}
\usepackage{color}
\definecolor{pred}{RGB}{255,90,90}
\definecolor{gt}{RGB}{90,173,90}
\definecolor{linegreen}{RGB}{0,204,0}
\definecolor{brightube}{rgb}{0.82, 0.62, 0.91}
\definecolor{byzantine}{rgb}{0.74, 0.2, 0.64}
\definecolor{citrine}{rgb}{0.89, 0.82, 0.04}

\newcommand{\good}[1]{{\color{red}#1}}
\newcommand{\linegreen}[1]{{\color{linegreen}#1}}

\AtBeginDocument{%
  }

\setcopyright{acmlicensed}
\copyrightyear{2024}
\acmYear{2024}
\acmDOI{10.1145/3664647.3681405}

\acmConference[MM '24]{Proceedings of the 32nd ACM International Conference on Multimedia}{October 28-November 1, 2024}{Melbourne, VIC, Australia}
\acmISBN{979-8-4007-0686-8/24/10}

\begin{document}

\title{\underline{SM$^4$}Depth: \underline{S}eamless \underline{M}onocular \underline{M}etric Depth Estimation across \underline{M}ultiple Cameras and Scenes by One \underline{M}odel}


\author{Yihao Liu}
\authornote{Both authors contributed equally to this research.}
\email{l1h_l1h_l1h@163.com}
\affiliation{
\institution{Beijing University of Posts and Telecommunications}
\country{China}}

\author{Feng Xue}
\authornotemark[1]
\email{feng.xue@unitn.it}
\affiliation{%
\institution{University of Trento}
\country{Italy}
}

\author{Anlong Ming}
\authornote{Corresponding author.}
\email{mal@bupt.edu.cn}
\affiliation{\institution{Beijing University of Posts and Telecommunications}
\country{China}}

\author{Mingshuai Zhao}
\email{mingshuai_z@bupt.edu.cn}
\affiliation{\institution{Beijing University of Posts and Telecommunications}
\country{China}}

\author{Huadong Ma}
\email{mhd@bupt.edu.cn}
\affiliation{\institution{Beijing University of Posts and Telecommunications}
\country{China}}

\author{Nicu Sebe}
\email{niculae.sebe@unitn.it}
\affiliation{\institution{University of Trento}
\country{Italy}}

\renewcommand{\shortauthors}{Liu et al.}

\begin{abstract}
In the last year, universal monocular metric depth estimation (universal MMDE) has gained considerable attention,
serving as the foundation model for various multimedia tasks,
such as video and image editing.
Nonetheless, current approaches face challenges in maintaining consistent accuracy across diverse scenes without scene-specific parameters and pre-training,
hindering the practicality of MMDE.
Furthermore, these methods rely on extensive datasets comprising millions,
if not tens of millions, of data for training,
leading to significant time and hardware expenses.
This paper presents SM$^4$Depth,
a model that seamlessly works for both indoor and outdoor scenes,
without needing extensive training data and GPU clusters.
Firstly,
to obtain consistent depth across diverse scenes,
we propose a novel metric scale modeling,
i.e., variation-based unnormalized depth bins.
It reduces the ambiguity of the conventional metric bins and enables better adaptation to large depth gaps of scenes during training.
Secondly,
we propose a ``divide and conquer" solution to reduce reliance on massive training data.
Instead of estimating directly from the vast solution space,
the metric bins are estimated from multiple solution sub-spaces to reduce complexity.
Additionally, we introduce an uncut depth dataset, BUPT Depth,
to evaluate the depth accuracy and consistency across various indoor and outdoor scenes.
Trained on a consumer-grade GPU using just 150K RGB-D pairs,
SM$^4$Depth achieves outstanding performance on the most never-before-seen datasets,
especially maintaining consistent accuracy across indoors and outdoors.
The code can be found \href{https://github.com/mRobotit/SM4Depth} {\textit{here}}.
\end{abstract}

\ccsdesc[500]{Computing methodologies~3D imaging}
\keywords{Seamless Monocular Metric Depth Estimation, Domain-aware Bin Estimation}

\begin{teaserfigure}
\centering
\includegraphics[width=1\linewidth]{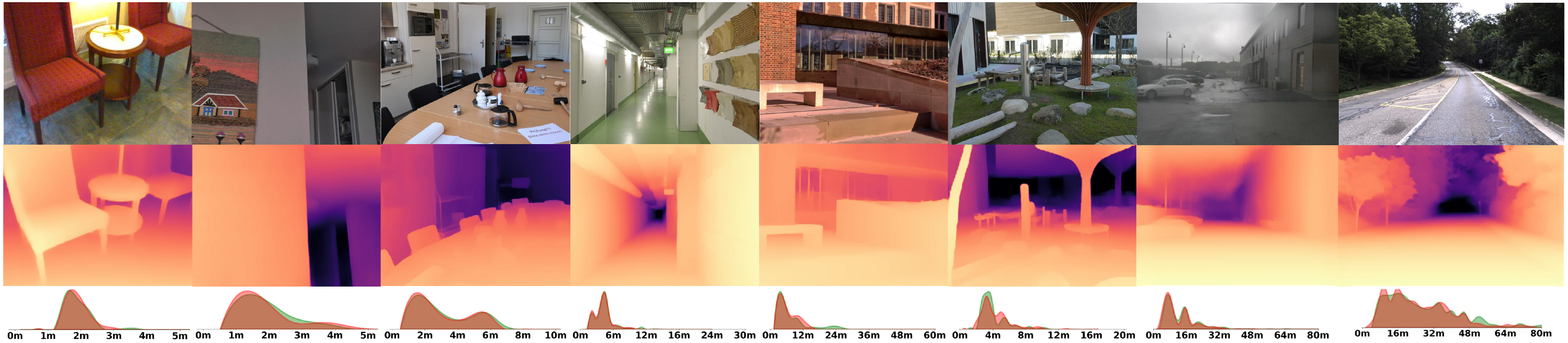}
\caption{Depth and distribution visualization of SM$^4$Depth that enables good generalization across multiple metric depth datasets captured by different sensors.
\textbf{Top:} input images.
\textbf{Middle:} depth prediction.
\textbf{Bottom:} distribution of the prediction (\textcolor{pred}{red}) and ground truth (\textcolor{gt}{green}).
\textbf{Six datasets:} SUN-RGBD\cite{sunrgbd}, DIODE\cite{diode}, iBims-1\cite{ibims}, ETH3D\cite{schoeps2017cvpr}, nuScenes-val\cite{Nuscenes2020}, and DDAD\cite{ddad}.}
\label{fig:intro_vis}
\end{teaserfigure}


 \maketitle

\section{Introduction}
\label{sec:intro}

Monocular depth estimation is a fundamental visual task with wide-ranging applications in the field of multimedia,
such as video editing \cite{chai2023stablevideo}
image editing \cite{10230895,zhang2023adding,10.1145/3532719.3543235},
3D generation/synthesis \cite{10.1145/3581783.3611800,10.1145/3581783.3612232,10.1145/3581783.3612033},
3D reconstruction \cite{10.1145/3581783.3612306,8490975},
and human pose estimation \cite{sun2022putting,8237635}.
In this community,
early research focused on MMDE
\cite{depthpr2021,10.1145/3474085.3475639,10.1145/3581783.3611751,10.1145/3474085.3475373,10.1145/3343031.3350930,10.1145/3240508.3240628,10.1145/3474085.3475386},
which were trained and tested only on specific datasets.
However,
they suffered from poor generalization when applied to unseen datasets,
which limited their applications in the real world.
To solve this issue,
much research shifted their focus to monocular relative depth estimation (MRDE) 
\cite{megadepth2018,BoostingDepth2022,ranftl2021vision} while disregarding the metric scale.
Leveraging diverse and easily accessible relative depth data,
these studies have achieved impressive performance,
enabling their application in scale-free tasks,
such as image editing \cite{Shufang,Chu} and image stylization \cite{ioannou2022depth,liu2017depth},
but not scale-sensitive applications,
e.g. virtual reality \cite{7500674,zhou2023occlusion,RuiLu-ICCV2019-OFNet}, 3D reconstruction \cite{cnnslam}, and even robot navigation \cite{xue_indoor_2023,tip,ICRA, chang2022fast,liang2023unknown}.

Beyond these approaches,
universal MMDE has recently gained prominence for its generalization capabilities,
marked by Depth Anything \cite{depthanything}, ZoeDepth \cite{ZoeDepth2023}, and Metric3D \cite{Metric3d2023}.
However, these methods still face challenges in the two aspects of MMDE:
\begin{enumerate}

\item \textbf{Inconsistent accuracy across scenes}:
The real world varies widely in depth,
ranging from $[1m,2m]$ (close-up scenes) to $[0.5m, 80m]$ (street scenes),
making models tend to focus on specific scenes and causing inconsistent accuracy across scenes.

\item \textbf{Heavy reliance on data amount}:
The reliance on massive training data (e.g. 8M metric depth data for Metric3D) remains due to the high complexity of determining a unique metric scale from a vast solution space of the natural scene.

\end{enumerate}

Aiming to address these issues,
we propose a \underline{S}eamless \underline{M}odel \underline{for} \underline{MMD}E across \underline{M}ultiple cameras and scenes (SM$^4$Depth for short).
First,
based on explicit modeling of metric scale,
we propose novel variation-based unnormalized depth bins which adaptively activate some parts rather than all of the fixed-length bins vector to describe the metric scale.
This reduces the bin ambiguity inherent in previous width-based bins,
and promotes the learning of widely different depth ranges in multiple scenes.
Regarding the second issue,
we propose a domain-aware bin estimation mechanism based on the ``divide and conquer'' idea,
which estimates metric bins from various solution sub-spaces, not the entire one, for reducing complexity.
\textbf{Divide}:
we divide the common depth range into several range domains (RDs) offline
and generate independent metric bins for each RD online.
\textbf{Conquer}:
we predict the RD that the input image belongs to and weightedly fuse all bins into a single one.
To verify the accuracy consistency across diverse scenes,
we propose BUPT Depth,
a seamless RGB-D dataset,
that consists of various indoor and outdoor scenes.
Owing to the design of SM$^4$Depth,
it performs superiously on the consistency of accuracy across diverse scenes,
which can be seen in Fig. \ref{fig:intro_vis}.
Notably,
it also achieves comparative performance compared with the state-of-the-art zero-shot MMDE methods but uses far fewer training samples (only 150K RGB-D pairs) and an affordable GPU (only single RTX 3090).

Our primary contributions are as follows:

\begin{enumerate}
\item 
SM$^4$Depth achieves consistent accuracy across diverse scenes using a single model,
eliminating the need for scene-specific parameters and pre-training.
This enhances the practicality of MMDE in real-world applications.
\item
This paper tackles the long-term unresolved issue of bin ambiguity using variation-based depth bins.
The proposed bins facilitate depth learning across scenes with significantly different depth ranges.
\item
Our domain-aware bin estimation mechanism reduces the reliance on massive training data in universal MMDE.
This enables SM$^4$Depth to achieve state-of-the-art accuracy with 150K RGB-D training pairs (only \textbf{0.02\%} amount used by previous methods) and a consumer-grade GPU.
\item
This paper presents the first no-clip RGB-D dataset. It is tailored to evaluate the consistent accuracy of MMDE methods \textbf{across diverse indoor and outdoor scenes}.
\end{enumerate}

\section{Related Work}
\noindent
\textbf{Monocular metric depth estimation} is a classic visual task,
in which determining metric scales is a crucial point,
and there are two paradigms.
Mainstream MMDE methods \cite{Laina2016,10.1145/3503161.3548381,10.1145/3474085.3475287,10.1145/3394171.3413706,10.1145/3503161.3549201,8486539,9852314,9669049} directly model this task as a pixel-wise regression problem (predicting continuous depth values in the real metric space),
where metric scales are implicitly encoded.
In contrast, since \cite{DORN_2018_CVPR},
several methods \cite{DANet2021,AdaBins2021,LocalBins2022} have defined this task as a classification problem.
Adabins \cite{AdaBins2021} explicitly encoded the metric scale into image-level depth bins.
We follow the latter paradigm as this paper focuses on recovering the metric scale.
However,
the same bin on two images with a large gap in depth range represents drastically different depths,
causing misleading back-propagation during training.
In this paper, we introduce variation-based bins to overcome this issue.

\noindent
\textbf{Zero-shot generation} has become a new trend in monocular depth estimation in recent years.
Early works \cite{megadepth2018,MiDaS2020,DPT2021} mainly achieved this goal by training with more accessible relative depth data.
Initially, Li \emph{et al.} \cite{megadepth2018} developed an MRDE pipeline on large-scale relative depth data.
Ranftl \emph{et al.} \cite{MiDaS2020} trained an MRDE model on five datasets and re-applied the training strategy to \cite{DPT2021}.
For high practicality,
the universal MMDE was first proposed 
in \cite{ZoeDepth2023} which combined relative depth and metric depth to achieve generalization.
Yin \emph{et al.} \cite{Metric3d2023} trained the model on 8M metric depth data for generalization.
Yang \emph{et al.} \cite{depthanything} proposed DepthAnything trained on 1M depth data and over 60M unlabeled data.
Piccinelli \emph{et al.} \cite{piccinelli2024unidepth} designed a camera-adaptive model, UniDepth,
and trained it on 300M metric depth images.
This reliance on numerous training data is due to the complexity of determining correct metric scales from diverse scenes.
Our approach aims to reduce this reliance.

\section{Problem Analysis and Countermeasures}

In this section,
we delve deep into the two issues of MMDE at the scene and data levels,
and provide specific solutions for each issue.

\subsection{Inconsistent Accuracy across Scenes}

Generally, real-world images exhibit vastly different depth ranges,
e.g. $[1m,2m]$ for indoor close-up and $[0.5m, 80m]$ for street scenes.
Such a large gap causes the model to overly concentrate on specific scenes instead of all scenes,
leading to 
\textbf{\textit{inconsistent accuracy across different scenes}}.
In this section,
we solve this issue by novel variance-based depth bins that bridge the gap of metric scale representation across scenes.
Before that,
we briefly review the conventional depth bin \cite{AdaBins2021} and outline its weakness.

\noindent
\textbf{Reviewing width-based depth bin and its weakness.}
Given the input image $I\!\in\!\mathbb{R}^{h\times w\times 3}$,
Adabins \cite{AdaBins2021} generates an $N$-channel probability map $P\in\mathbb{R}^{h\times w\times N}$ 
and a vector $c\in\mathbb{R}^{N\times1}$
representing the centers of $N$ depth bins discreted from the depth interval,
which are linearly combined
to obtain a metric depth map $D\in\mathbb{R}^{h\times w}$:
\begin{equation}
\setlength{\abovedisplayskip}{4pt}
\setlength{\belowdisplayskip}{4pt}
D(i) = \sum\nolimits_{n=1}^{N}c_n P_n(i)
\label{eq:cp}
\end{equation}
where $D(i)$ is the $i^{\text{th}}$ pixel's predicted depth,
and $P_n(i)$ denotes the probability for pixel $i$ that its depth is equal to the $n^\text{th}$ bin center $c_n$.
In Eq. \eqref{eq:cp},
the bin center $c_n$ is calculated by accumulating the width of each bin $b\in\mathbb{R}^{N\times 1}$:
\begin{equation}
\setlength{\abovedisplayskip}{4pt}
\setlength{\belowdisplayskip}{4pt}
\label{eq:center}
c_n=d_\text{min}+\left(d_\text{max}-d_\text{min}\right)(b_n/2+\sum\nolimits_{j=1}^{n-1}b_j)
\end{equation}
where $b_n=(b^{'}_{n}+\epsilon)/\sum_{i=1}^{N}(b^{'}_{i}+\epsilon)$ denotes 
the normalized width of the $n^\text{th}$ depth bin, with $\epsilon=10^{-3}$
and $b^{'}_{n}\in[0,+\infty)$ being the unnormalized width predicted 
through a feedforward neural network (FFN) with the ReLU activation function.
During training,
the bi-directional Chamfer loss \cite{ChamferLoss2017} is employed to enforce the small width $b^{'}$ 
within the interesting depth interval in the ground truth depth map $\mathbf{D}$:
\begin{equation}
\setlength{\abovedisplayskip}{4pt}
\setlength{\belowdisplayskip}{4pt}
\label{eq:chamfer}
\mathcal{L}_{bin}(c,\mathbf{D})=\sum\nolimits_{\mathbf{d}\in \mathbf{D}}\!\min_{c_n\in c}\!||\mathbf{d}-c_n||^2+\sum\nolimits_{c_n\in c}\!\min_{\mathbf{d}\in \mathbf{D}}\!||\mathbf{d}-c_n||^2
\end{equation}
where $\mathbf{d}$ is the pixel's correct depth.

\begin{figure}
\centering\includegraphics[width=1\linewidth]{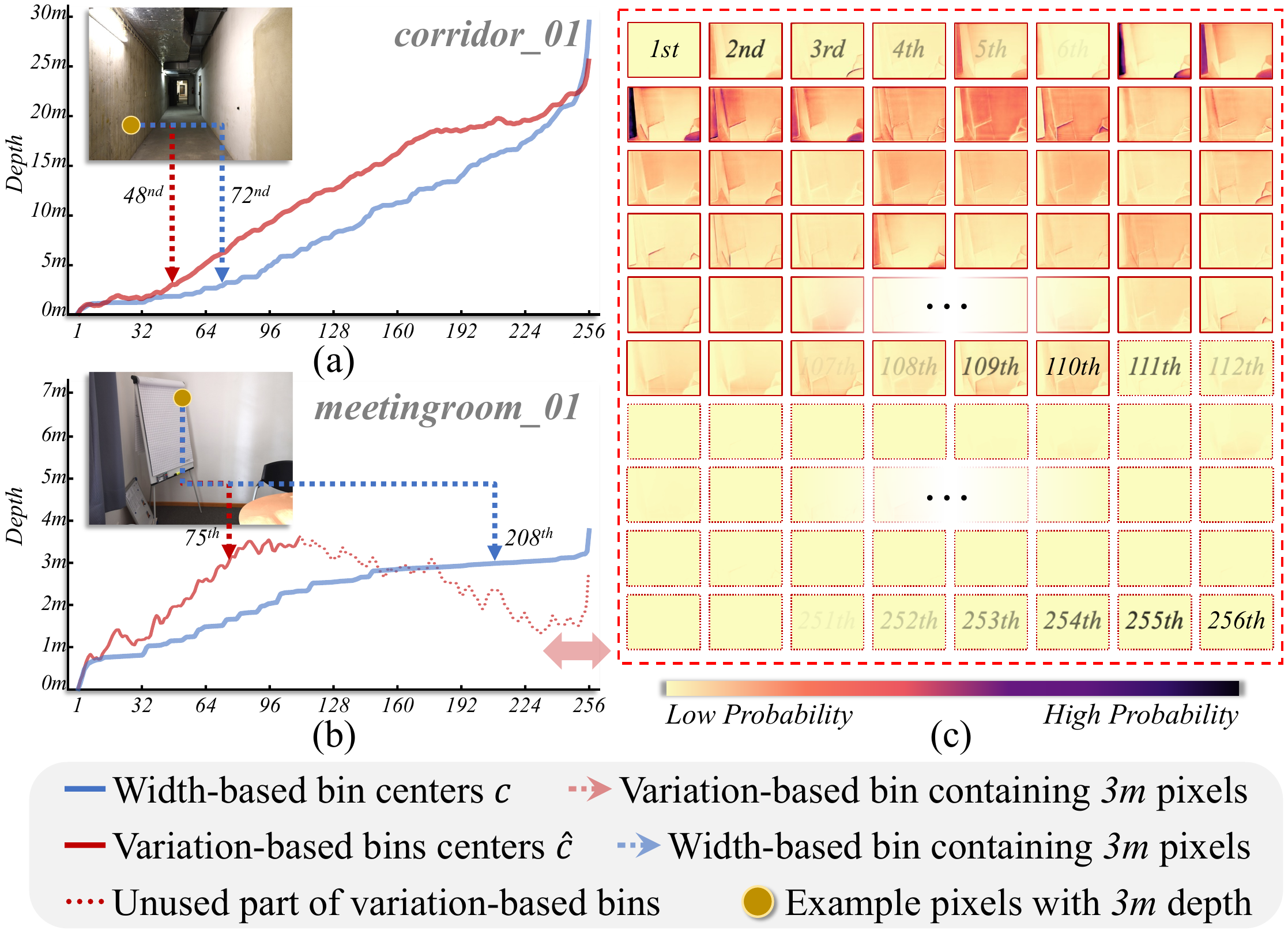}
\caption{
Two bin-center curves on (a) a distant-view image with range $[0m,30m]$ and (b) a close-up one with range $[0m,3.5m]$ from iBims-1 \cite{ibims}.
(c) represents probability maps $P$ corresponding to the red curve in (b).
}
\label{fig:ca}
\vspace{-10pt}
\end{figure}

In natural scenes,
the depth range of images varies significantly,
yet all images must represent metric scales using the $N$ bins.
This leads to great variations in the metric depth represented by a single bin,
a phenomenon we call ``\textit{bin ambiguity}''.
Taking Fig. \ref{fig:ca} as an example,
a 3$m$ pixel would be classified in the 208$^\text{th}$ bin for an image with a range of $[0, 3.5m]$ (Fig. \ref{fig:ca}(b)),
but in the 72$^{th}$ bin for an image with a range of $[0, 30m]$ (Fig. \ref{fig:ca}(a)).
Such an excessive gap would confuse the metric meaning of the probability map's channels
and lead to the back-propagation of misleading signals during training.

\noindent
\textbf{Depth variation-based bins for consistent accuracy.}
\label{sec:vb}
The intuitive idea is to use the front portion of bins for short-range images and the entire bins for long-range images.
To achieve this,
we propose the variation-based unnormalized depth bins.
Unlike the conventional bin $b^{'}_n$,
we use only an FFN without ReLU activation.
In this way,
the FFN outputs variations that allow negative values,
denoted as $\hat{b}^{'}\in\mathbb{R}^{N\times1}$.
Then,
the bin center $c$ in Eq. \eqref{eq:center} is re-formulated to an unnormalized bin center $\hat{c}$,
which is no longer limited by the depth range of specific datasets
(e.g., $[0m,10m]$ for NYUDv2 and $[0m,80m]$ for KITTI):
\begin{equation}
\setlength{\abovedisplayskip}{4pt}
\setlength{\belowdisplayskip}{4pt}
\hat{c}_n=\epsilon+\hat{b}_n^{'}/2+\sum\nolimits_{j=1}^{n-1}\hat{b}_j^{'}
\label{eq:ubc}
\end{equation}

\textbf{Analysis:}
Since the depth variations $\hat{b}^{'}$ are allowed to be negative,
the bin center can reach the maximum depth on some intermediate bin $\hat{c}_\mathbf{n} (\mathbf{n}<N)$ in short-range images,
not necessarily the last bin center $\hat{c}_N$.
Thus, all pixels can be fully expressed by the front bins $\{\hat{c}_n| n\in[1,\mathbf{n}]\}$,
and do not have to involve the latter bins $\{\hat{c}_n| n\in(\mathbf{n},N]\}$.
As indicated by the red lines in Fig. \ref{fig:ca},
for the close-up image,
the bin center reaches the maximum depth at the 110$^{th}$ bin and then continues to decrease.
While the later bins $\{\hat{c}_n| n\in(110,N]\}$ correspond to some probability channels close to zero $\{P_n| n\in(110,N]\}$,
which can be observed through the visualization of part channels in Fig. \ref{fig:ca} (c).
This illustrates that these later bins are not actually used.

\begin{figure}
\centering\includegraphics[width=1\linewidth]{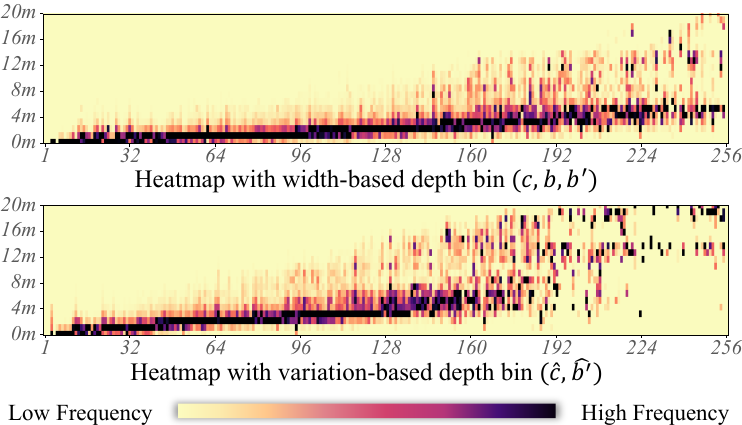}
\caption{\textbf{The heatmaps show the frequency of depth values occurring in each depth bin},
which are obtained from iBims-1 \cite{ibims}.
If a square $(\mathcal{X},\mathcal{Y})$ appears darker,
it indicates that the depth value $\mathcal{Y}$ mainly occurs 
within the $\mathcal{X}^\text{th}$ depth bin.}
\label{fig:heatmap}
\vspace{-10pt}
\end{figure}

Fig. \ref{fig:heatmap} presents additional statistics information,
i.e., the frequency of depth values occurring in each depth bin.
For the width-based depth bin $(c,b,b^{'})$,
depths below $4m$ occur most frequently across all bins.
Conversely, variation-based depth bin $(\hat{c},\hat{b}^{'})$
exhibits larger depths in the latter bins.
This means that the depth values represented by each bin center are pulled apart on the level of the entire dataset,
suppressing the bin ambiguity.

\subsection{Reliance on Massive Training Data}
\label{sec:sq}

\noindent
\textbf{Reason behind the reliance.}
Practical applications differ from specific datasets in that images are taken from various camera angles and innumerable scenes.
Due to the diverse nature of appearance,
mapping from visual cues to a wide range of depth values becomes highly intricate and cannot be exhaustively presented.
Consequently, 
determining metric depth bins entails exploring a vast solution space, 
which necessitates greater attention to reducing its complexity. 
However,
previous works have overlooked this crucial aspect by directly making predictions (e.g., \cite{ZoeDepth2023} solves for metric bins and \cite{Metric3d2023} predicts metric depth) 
from the entire solution space, inevitably requiring massive training data.
To address this issue,
we first divide the whole solution space into several sub-spaces.
Then, a ``divide and conquer'' method is proposed to generate metric bins in each sub-space and predict the best metric bins for the input.

\vspace{5pt}
\noindent
\textbf{Stage 1: Online depth range domain generation.}
To divide the solution space into sub-spaces,
the previous approaches group all images according to semantic categories \cite{ZoeDepth2023,BinsFormer2022}.
However, a large gap in depth range may exist within one scene.
Differently,
we group all training images according to the depth range that better constrains the perspective and scene from which the image is taken.

According to \cite{DORN_2018_CVPR},
the amount of information for depth estimation decreases as the depth value increases.
Thus, we employ a space-increasing strategy to gain more image groups (named range domain, RD) when the depth value is smaller.
Assuming that the depth range is $[Z_\text{min}, Z_\text{max}]$ and there are $K$ RDs,
the $k^\text{th}$ RD can be formulated as:
\begin{equation}
\setlength{\abovedisplayskip}{4pt}
\setlength{\belowdisplayskip}{4pt}
RD_k= \Big[ Z_\text{min}, Z_\text{min} + \sum\nolimits_{i=1}^{k}{\frac{2i(Z_\text{max} - Z_\text{min})}{K (1+K)}}\Big]
\end{equation}
We further visualize the RDs in the supplementary material \href{https://pan.baidu.com/s/1CCpvQ4rbVkmagVntP_XFLQ?pwd=g1ct}{\textit{here}}.

\vspace{5pt}
\noindent
\textbf{Stage 2: Online domain-aware bin estimation design.}
We design a domain-aware bin estimation mechanism that generates metric bins for each RD and finds the best-matching metric bins, following the ``divide and conquer'' idea in two steps.

\label{sec:dbed}

\textbf{The ``Divide'' step} aims to discretize each depth interval $RD_k$ into $N$ bins.
Specifically, given the deep feature of the input image,
we leverage a transformer encoder
to learn the relationship between the deep feature
and $K$ preset learnable 1-D embeddings (called bin queries).
The output embeddings of these queries are fed into an FFN 
to generate $K$ depth variation vectors $\{\hat{b}^{'[k]}|k\!\!\in\!\![1,K]\}$,
and calculate the bin center vectors $\{\hat{c}^{[k]}|k\!\in\![1,K]\}$ using Eq. \eqref{eq:ubc}. 
To illustrate our idea,
we compare three possible design choices:
\begin{itemize}
    \item \textbf{$1$ Query + $K$ FFNs}: Using $K$ FFNs to process the output of only one query.
    \item \textbf{$K$ Queries + $K$ FFNs}: Using $K$ FFNs to process the outputs of $K$ queries in a one-to-one way.
    \item \textbf{$K$ Queries + $1$ FFN (Ours)}: Using only one FFN to process the outputs of $K$ queries.
\end{itemize}
The first two both employ $K$ FFNs.
Thus, each FFN only learns the knowledge of a single RD during training,
which leads to drastically different outputs of these FFNs and 
makes them sensitive to input noise.
The last design is recommended as the best choice 
and the experiments (in Sec.\ref{sec:detail analysis}) verify its superiority over other options.

\textbf{The ``Conquer'' step} aims to estimate the correct RD for the input image and determine the best-matching metric bins.
Specifically,
we preset an additional 1-D embedding (called domain query) 
alongside the bin queries.
Its corresponding output is then fed into a classification head (CLS) to generate the probability that the input image belongs to each RD,
denoted as $\{y_{k} \in [0,1] | k\in[1,K] \}$.

Subsequently,
considering the possibility of images being positioned near the decision boundary of RD classification,
we do not select the top-scoring metric bin
but instead combine all bin center vectors $\hat{c}^{[k]}$ to a single one by using the RD probabilities $\{y_{k}| k\in[1,K] \}$ as weights:
\begin{equation}
\setlength{\abovedisplayskip}{4pt}
\setlength{\belowdisplayskip}{4pt}
\mathbf{c} = \sum\nolimits_{k\in[1,K]} \hat{c}^{[k]} y_{k}
\end{equation}
where $\mathbf{c}$ is the final bin center vector.

\begin{figure}
\centering\includegraphics[width=1\linewidth]{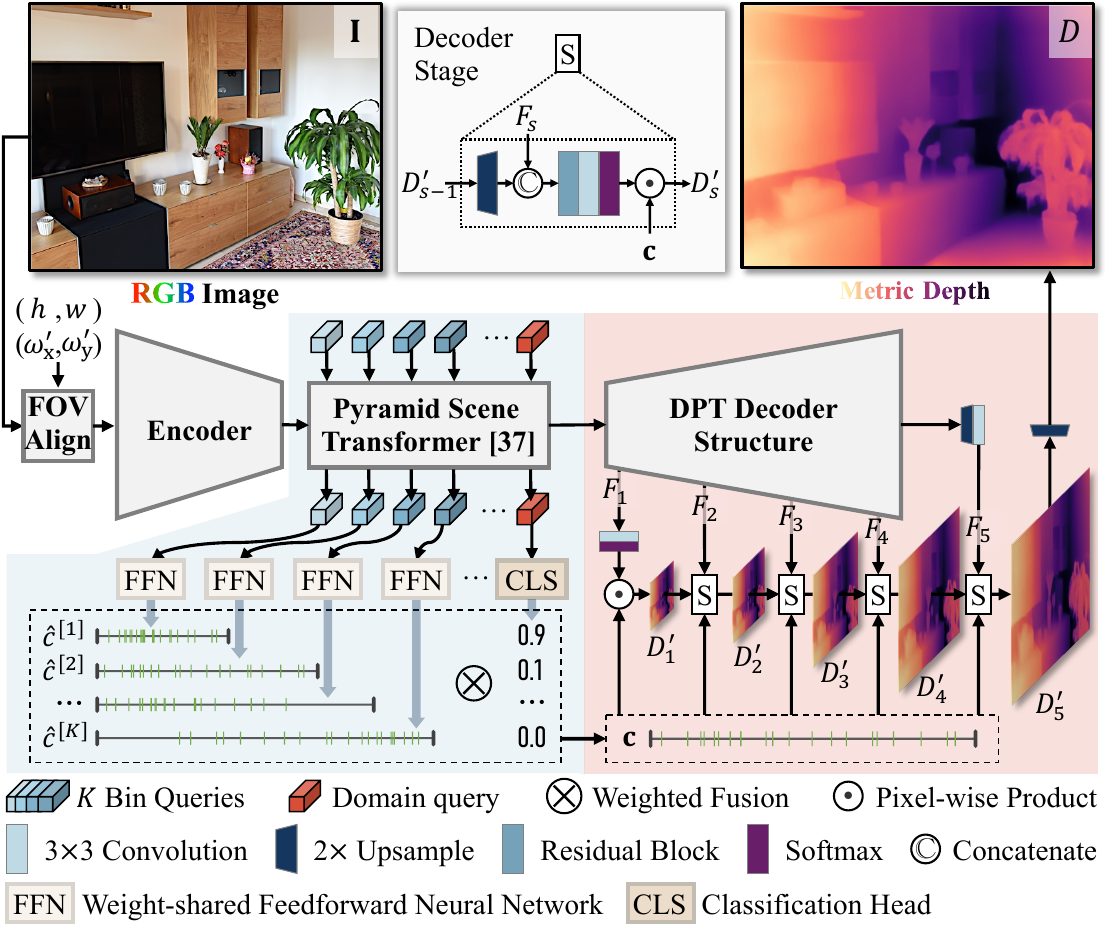}
\caption{SM$^4$Depth Pipeline containing the domain-aware bin estimation (\textcolor{blue}{blue} mask) and the HSC-decoder (\textcolor{red}{red} mask).}
\label{fig:architecture}
\vspace{-10pt}
\end{figure}

\begin{figure*}
\centering
\includegraphics[width=1\linewidth]{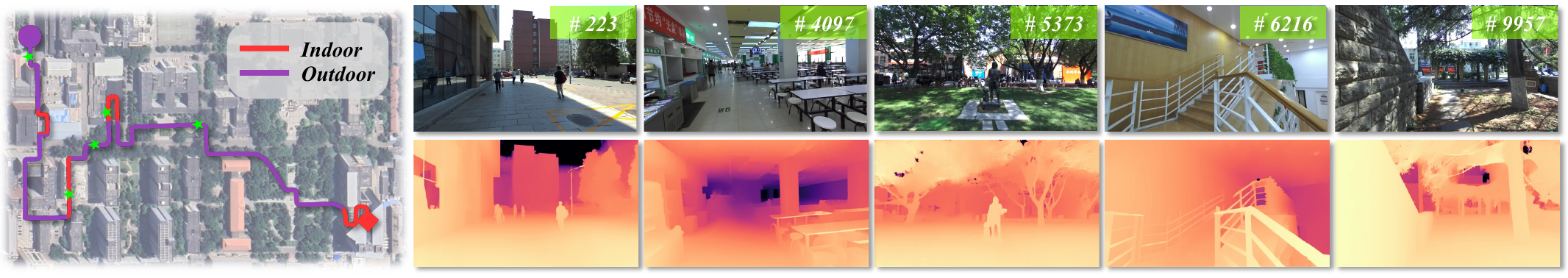}
\caption{Top view of the scene where we collected the BUPT Depth dataset.
Red lines indicate indoor scenes
and purple lines indicate outdoor scenes.
We give five images and their ground truth depth calculated by CREStereo \cite{CRESTEREO2022} as examples.}
\label{fig:bupt}
\vspace{-10pt}
\end{figure*}

\section{Architecture of SM$^4$Depth}

\subsection{Pipeline.}
Fig. \ref{fig:architecture} illustrates the structure of our network.
Given an RGB image $\mathbf{I} \!\in\!\mathbb{R}^{\mathbf{h}\times\mathbf{w}\times3}$,
we first pre-process $\mathbf{I}$ to obtain a new image $I \!\in\! \mathbb{R}^{h\times w\times3}$ with a unified field of view (FOV) (see Sec.\ref{sec:pre-processing}).
Then,
we extract the deep feature from $I$ by an encoder.
Next,
a pyramid scene transformer (PST) \cite{DANet2021} is positioned between the encoder and decoder.
It consists of three parallel transformer encoders with inputs of different patch sizes, respectively.
We employ the transformer encoder with the smallest patch size to process all queries.
Based on the mechanism in Sec.\ref{sec:dbed},
we obtain the bin center vector $\mathbf{c}$ of image $I$.
Finally,
we design a decoder with hierarchical scale constraints (HSC-Decoder) to anchor the metric scale in multiple resolutions and output the metric depth map $D$ (see Sec.\ref{sec:hscdecoder}).

\subsection{FOV alignment Pre-Processing.}
\label{sec:pre-processing}
According to \cite{Metric3d2023},
eliminating ``metric ambiguity'' is the key to achieving universal MMDE.
Therefore,
we pre-process all input images to align their field of view (FOV).
Given an input image $\mathbf{I}$ with focal length $(\mathbf{f_x},\mathbf{f_y})$,
we first preset the input resolution of network as $(h,w)$ and define the target FOV as $(\omega^{'}_\mathbf{x},\omega^{'}_\mathbf{y})$ 
in radians.
Then,
a rectangular region $I^{'} \!\in\!\mathbb{R}^{h^{'}\!\times w^{'}\!\times3}$ on $\mathbf{I}$ equivalent to the target FOV $(\omega^{'}_\mathbf{x},\omega^{'}_\mathbf{y})$ is calculated by the FOV equation:
\begin{equation}
\setlength{\abovedisplayskip}{4pt}
\setlength{\belowdisplayskip}{4pt}
\!\!w^{'} = 2\mathbf{f_x}\tan(\omega^{'}_\mathbf{x}/2)\,\, , \,\,h^{'} = 2\mathbf{f_y}\tan(\omega^{'}_\mathbf{y}/2)
\end{equation}
Next, we crop this region $I^{'}$ from the original image $\mathbf{I}$,
and fill the pixels beyond $\mathbf{I}$ with 255.
Finally, the region $I^{'}$ is resized to the target resolution $(h,w)$ to generate a new image $I$ as input of the network.
Note that this FOV alignment is similar to the CSTM\_image \cite{Metric3d2023} in result,
but without maintaining a canonical camera space.
Thus it is more straightforward.

\begin{table*}[t]
\scriptsize
\centering
\resizebox{\linewidth}{!}{
\begin{tabular}{l|l|c|c|c|cccccc|cccccc}
\toprule
\multirow{2}{*}{\textbf{Method}} & \multirow{2}{*}{\textbf{Backone}} & \multirow{2}{*}{\makecell[c]{\textbf{Training}\\\textbf{Pairs}}} & \multirow{2}{*}{\textbf{FLOPs}} & \multirow{2}{*}{\textbf{Params}} & \multicolumn{6}{c|}{ \textbf{BUPT Depth - Ground Truth gained by ZED2}}& \multicolumn{6}{c}{ \textbf{BUPT Depth - Ground Truth gained by CREStereo}} \\
& & & & & $\mathbf{\delta_1 \uparrow}$ & $\mathbf{\delta_2 \uparrow}$ & $\mathbf{\delta_3 \uparrow}$ &\textbf{ REL$\downarrow$} & \textbf{RMSE$\downarrow$} & \textbf{log10$\downarrow$} & $\mathbf{\delta_1 \uparrow}$ & $\mathbf{\delta_2 \uparrow}$ & $\mathbf{\delta_3 \uparrow}$ & \textbf{REL$\downarrow$} & \textbf{RMSE$\downarrow$} & \textbf{log10$\downarrow$}\\
\midrule
ZoeD-NK \cite{ZoeDepth2023} & BEiT-L & - & 283G & 267M & \underline{0.314} & \underline{0.485} & 0.576 & {1.006} & \underline{9.758} & {0.263} & \underline{0.365} & \underline{0.505} & {0.579} & \underline{0.974} & \underline{9.377} & \underline{0.252}\\
Metric3D \cite{Metric3d2023} & CNXT-L & over 8M & 569G & 203M & 0.155 & 0.318 & 0.417 & 2.446 & 23.794 & 0.427 & 0.293 & 0.402 & 0.470 & 1.816 & 22.736 & 0.352\\
DepthAnything-NK \cite{depthanything} & ViT-L & over 61M & 451.9G & 335.3M & 0.193 & 0.3623 & \textbf{1.000} & 1.1362 & 7.9838 & 0.2978 & 0.221&0.3742&\textbf{1.000}&1.120&7.833&0.292\\
UniDepth \cite{piccinelli2024unidepth} & ViT-L & roughly 3M & -  & 347M & 0.311 & 0.598 & \textbf{1.000} & \underline{0.6559} & 8.399 & \underline{0.1933} & 0.109 & 0.396 & \textbf{1.000} & 0.983 & 9.422 & 0.264 \\
\textbf{SM$^4$Depth} & \textbf{Swin-B} & \textbf{0.015M}& \textbf{105G} & \textbf{110M} & \textbf{0.536} & \textbf{0.805} & \underline{0.924} & \textbf{0.295} & \textbf{3.440} & \textbf{0.118}& \textbf{0.629} & \textbf{0.875} & \underline{0.966} & \textbf{0.241} & \textbf{2.888} & \textbf{0.094} \\
\bottomrule
\end{tabular}}
\caption{\textbf{Quantitative results on BUPT Depth.} 
Comparisons are conducted with the ground truths of ZED and CREStereo, respectively.
The best results are in bold, and the second-best ones are underlined.
- indicates unknown numbers.}
\label{tab:quantitative res bupt depth}
\vspace{-15pt}
\end{table*}

\subsection{Decoder with hierarchical scale constraints.}
\label{sec:hscdecoder}
Our decoder draws inspiration from the refinement decoder structures \cite{BTS2019,SARPN2019},
but the divergence lies in scale constraints on the metric depth at each stage.
As shown in Fig. \ref{fig:architecture}, 
taking the PST's output, denoted as $F_1$, as input,
we employ the DPT’s decoder \cite{DPT2021} to gradually recover the resolution of features,
denoted as $F_s$ with a size $\frac{h}{2^{(6-s)}}\times\frac{w}{2^{(6-s)}}$,
where $s\in\{1,2,3,4,5\}$ is the stage number.
In the first stage,
$F_1$ is compressed into $N$-channel and then multiplied pixel-wisely with $\mathbf{c}$ by Eq. \eqref{eq:cp},
generating a low-resolution depth map $D^{'}_1\in\mathbb{R}^{\frac{h}{32}\times\frac{w}{32}}$.
In the following $s^\text{th}$ stage,
the depth map of the former stage $D^{'}_{s-1}$ is upsampled 
and fused with feature $F_s$ by a residual convolution block \cite{SARPN2019}.
Then we linearly combine the fused feature and the bin centers $\mathbf{c}$ 
to generate the depth map $D^{'}_s\in\mathbb{R}^{\frac{h}{2^{(6-s)}}\times\frac{w}{2^{(6-s)}}}$.
In this way,
the depth map of the last stage $D^{'}_5\in\mathbb{R}^{\frac{h}{2}\times\frac{w}{2}}$ is obtained.
Compared to the previous refinement decoder \cite{SARPN2019},
the HSC-Decoder incorporates the metric bins into each stage to progressively refine the depth range,
thus performing better in recovering the depth range.
The loss functions are further described in the supplementary material \href{https://pan.baidu.com/s/1CCpvQ4rbVkmagVntP_XFLQ?pwd=g1ct}{\textit{here}}.

\section{Uncut RGBD dataset: BUPT Depth}
BUPT Depth (see in Fig. \ref{fig:bupt}) is proposed to evaluate consistency in accuracy across indoor and outdoor scenes,
including streets, canteen, classroom, and lounges, etc.
This dataset shows a variety of lighting, scenes, and viewing angles, making depth estimation challenging.
It consists of 14,932 continuous RGB-D frames captured in BUPT by ZED2.
In addition to the outputs of ZED2, we provide the re-generated depth maps from CREStereo \cite{CRESTEREO2022} 
and the sky segmentation from ViT-Adapter \cite{ViT-Adapter2022}.
The color and depth streams are captured with intrinsics of $1091.517$
and a baseline of $120.034 \mathrm{mm}$.

\begin{figure*}
\centering
\includegraphics[width=1\linewidth]{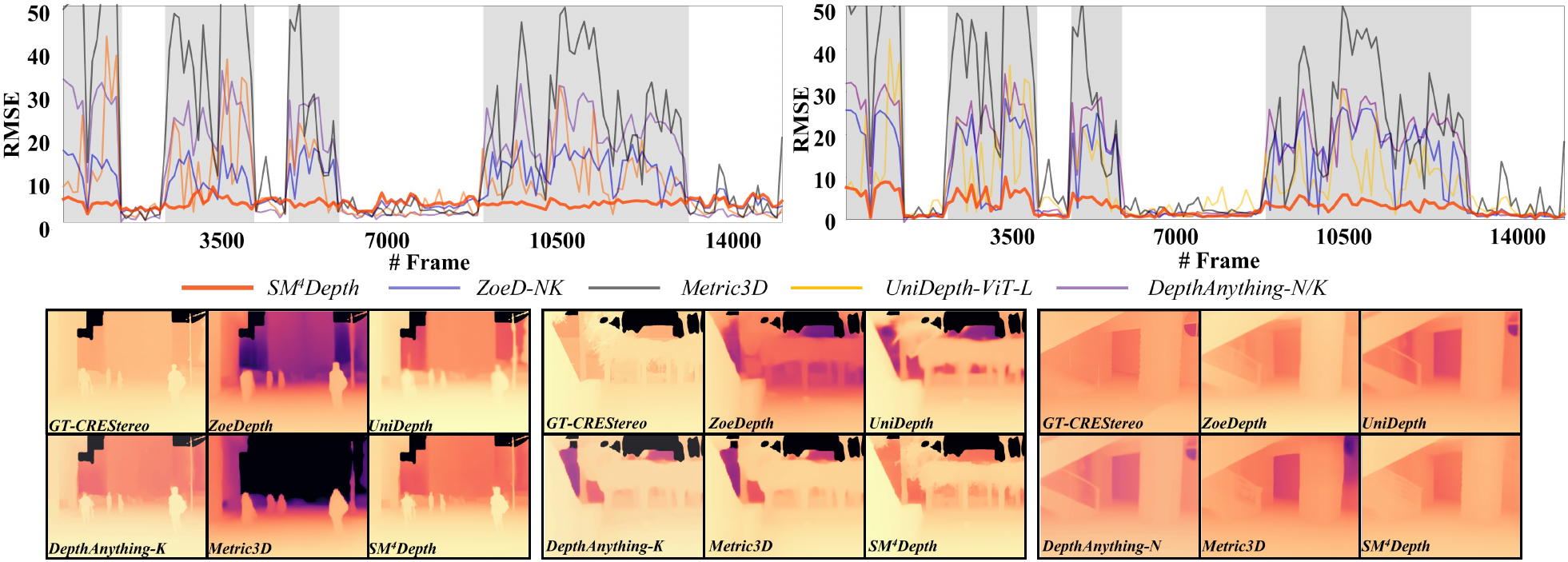}
\caption{
\textbf{RMSE per frame of SM$^4$Depth (\textcolor{orange}{orange}), Metirc3D (\textcolor{gray}{gray}), ZoeDepth-NK (\textcolor{blue}{blue}), UniDepth (\textcolor{citrine}{yellow}), DepthAnything-NK(\textcolor{purple}{purple}) on BUPT Depth.}
We use the stereo depth of ZED2 (\textbf{the first chart}) and CREStereo\cite{CRESTEREO2022} (\textbf{the second chart}) as ground truth, respectively. 
Gray indicates outdoor frames,
and white indicates indoor frames.}
\label{fig:zed scale shift}
\vspace{-5pt}
\end{figure*}

\section{Experiments}
\label{sec:exps}
\subsection{Experimental setting}
\noindent
\textbf{Datasets:}
For training,
we randomly sample RGB-Depth pairs from various datasets.
Specifically, we sample
24K pairs from ScanNet \cite{scannet},
15K pairs from Hypersim \cite{Hypersim},
51K pairs from DIML \cite{diml},
36K pairs from UASOL \cite{uasol},
14K pairs from ApolloScape \cite{apolloscape},
and 11K pairs from CityScapes \cite{cityscapes}.
During training, NYUD \cite{nyud} and KITTI \cite{kitti} are used for validation.
In addition,
we apply the same pre-processing steps to the training data as \cite{MiDaS2020,ZoeDepth2023},
elaborated in the supplementary material \href{https://pan.baidu.com/s/1CCpvQ4rbVkmagVntP_XFLQ?pwd=g1ct}{\textit{here}}.
During testing,
we employ eight datasets not seen during training: 
SUN RGB-D \cite{sunrgbd}, iBims-1 \cite{ibims}, ETH3D Indoor/Outdoor \cite{eth3d}, DIODE Indoor/Outdoor \cite{diode},
nuScenes-val \cite{Nuscenes2020}, and DDAD \cite{ddad}. 
Note that we remove the test set of NYUD from SUN RGB-D for a fair comparison.

\noindent
\textbf{Metrics:}
We employ four metrics \cite{AdaBins2021} for evaluation: 
the accuracy under threshold ($\delta_k < 1.25^k, k=1,2,3$),
the absolute relative error (REL), and the root mean squared error (RMSE).
In addition, we use the relative improvement across datasets (mRI$_\eta$) and metrics (mRI$_\theta$) in \cite{ZoeDepth2023}.
During the evaluation,
the final output is obtained by averaging the predictions for an image and its mirror image.
In addition,
the final output is upsampled to match the original image size, 
and all metrics are computed within the same FOV.

\noindent
\textbf{Implementation Details:}
SM$^4$Depth employs the Swin Transformer Base as the backbone, 
and runs on a single NVIDIA RTX 3090 GPU. 
The network is trained by the Adam optimizer with parameters $(\beta_1, \beta_2)=(0.9, 0.999)$. 
The training runs for 20 epochs with a batch size of 10.
The initial learning rate is set to $2\times 10^{-5}$ and gradually reduced to $2\times 10^{-6}$.
Note that, an over-large fixed FOV would cause too large invalid area in the small FOV dataset, making the network underfitting.
We empirically set the fixed FOV to $(\omega^{'}_\mathbf{x},\omega^{'}_\mathbf{y})\!=\!(58^{\circ},45^{\circ})$ and the fixed resolution to $(w,h)\!=\!(564,424)$.

\subsection{Result on BUPT Depth}
Table \ref{tab:quantitative res bupt depth} presents a quantitative comparison between SM$^4$Depth and state-of-the-art methods,
categorized based on different ground truth.
SM$^4$Depth achieves superior performance in most metrics when utilizing ground truth from either ZED2 or CREStereo \cite{CRESTEREO2022}.
Notably, SM$^4$Depth is trained on a smaller dataset of 0.015M pairs (only 0.02\%$\sim$5\% of the data used by previous methods),
with FLOPs and parameter count typically half that of prior methods.

Fig. \ref{fig:zed scale shift} provides a more intuitive comparison between SM$^4$Depth and state-of-the-art methods from a time-series perspective,
visualizing the RMSE of each frame using two line graphs.
SM$^4$Depth (orange) achieves a relatively high accuracy regardless of indoor or outdoor scenes on both two ground truth.
In contrast, other methods fail to achieve high accuracy simultaneously indoors and outdoors.
Specifically, ZoeD-NK, Metric3D, and UniDepth show fluctuating accuracy in outdoor scenes,
with some even exceeding over 50 at times.
And the accuracy of SM$^4$Depth varies more smoothly.
In the first two examples we presented in Fig. \ref{fig:zed scale shift},
both SM$^4$Depth and UniDepth obtain a more accurate metric scale than others,
but SM$^4$Depth provides sharper outputs,
even enabling the wooden fence in frame 9957 to be discernible.
In the indoor scene,
SM$^4$Depth produces clearer outlines than other methods.

\begin{table*}[t!]
\centering
\resizebox{\linewidth}{!}{
\begin{tabular}{cl|cccc|cccc|cccc|cccc}
\toprule
\multirow{2}{*}{\textbf{Categories}} & \multirow{2}{*}{\textbf{Method}} & \multicolumn{4}{c|}{\textbf{SUN RGB-D}} &
\multicolumn{4}{c|}{\textbf{iBims-1 Benchmark}} & \multicolumn{4}{c|}{\textbf{ETH3D Indoor}} &
\multicolumn{4}{c}{\textbf{DIODE Indoor}} 
\\ & 
& ~$\mathbf{\delta_1 \uparrow}$~ & \textbf{REL~$\downarrow$} & \textbf{RMSE~$\downarrow$} & \textbf{mRI$_\theta\uparrow$}
& ~$\mathbf{\delta_1 \uparrow}$~ & \textbf{REL~$\downarrow$} & \textbf{RMSE~$\downarrow$} & \textbf{mRI$_\theta\uparrow$} 
& ~$\mathbf{\delta_1 \uparrow}$~ & \textbf{REL~$\downarrow$} & \textbf{RMSE~$\downarrow$} & \textbf{mRI$_\theta\uparrow$}
& ~$\mathbf{\delta_1 \uparrow}$~ & \textbf{REL~$\downarrow$} & \textbf{RMSE~$\downarrow$} & \textbf{mRI$_\theta\uparrow$} 
\\ \midrule
\multirow{4}{*}{\textbf{Full-shot}} & BTS \cite{BTS2019} & 0.718 & 0.181 & 0.533 & -31.45\% & 0.536 & 0.233 & 1.059 & -32.82\% & 0.360 & 0.324 & 2.210 & -18.73\% & 0.208 & 0.419 & 2.382 & -34.57\%\\ 
& AdaBins \cite{AdaBins2021}                             & 0.751 & 0.167 & 0.493 & -23.07\% & 0.548 & 0.216 & 1.078 & -29.56\% & 0.283 & 0.361 & 2.347 & -31.23\% & 0.173 & 0.442 & 2.450 & -40.95\%\\ 
& NeWCRFs \cite{NeWCRFs2022}                             & 0.779 & 0.159 & 0.437 & -14.67\% & 0.543 & 0.209 & 1.031 & -26.43\% & 0.452 & 0.268 & 1.874 & 0.76\% & 0.183 & 0.402 & 2.307 & -33.51\% \\
& MIM \cite{MIM2022}                                     & 0.844 & 0.147 & 0.341 & 0.00\% & 0.717 & 0.163 & 0.813 & 0.00\%  & 0.453 & 0.287 & 1.800 & 0.00\% & 0.416 & 0.317 & 1.960 & 0.00\%\\ 
\midrule
\multirow{7}{*}{\makecell{\textbf{Zero-shot}}} 
& ZoeD-N \cite{ZoeDepth2023}                             & 0.850 & 0.125 & 0.357 & 3.66\% & 0.652 & 0.171 & 0.883 & -7.53\% & 0.388 & 0.275 & 1.678 & -1.13\% & 0.376 & 0.331 & 2.198 & -8.72\%\\ 
& DepthAnything-N \cite{depthanything}                   & \underline{0.897} & \underline{0.107} & \underline{0.272} & \underline{17.90\%} & 0.716 & \underline{0.150} & 0.726 & 6.17 \%& 0.438 & 0.250 & 1.663 & 5.73 & 0.277 & 0.339 & 2.113 & -16.05 \\
& \textbf{SM$^4$Depth-N}                                 & 0.874 & 0.121 & 0.303 & 10.80\%  & 0.715 & 0.162 & 0.801 & 0.60\%  & 0.486 & \underline{0.249} & 1.662 & \underline{9.40\%}  & 0.418 & \underline{0.298} & 1.790 & \underline{5.05\%} \\ 
\cmidrule(lr){2-18}
& ZoeD-NK \cite{ZoeDepth2023}                            & 0.841 & 0.129 & 0.367 & 1.42\% & 0.610 & 0.189 & 0.952 & -15.99\% & 0.353 & 0.280 & 1.691 & -4.53\%  & 0.386 & 0.335 & 2.211 & -8.57\% \\ 
& Metric3D\cite{Metric3d2023}                            & 0.033 & 2.631 & 5.633 & × & \textbf{0.818} & 0.158 & \textbf{0.582} & \textbf{15.18\%} & \textbf{0.536} & 0.335 & \underline{1.550} & 5.16\% & \underline{0.505} & 0.427 & \underline{1.687} & 0.21\% \\
& UniDepth-ViT-L \cite{piccinelli2024unidepth}           & \textbf{0.953} & \textbf{0.089} & \textbf{0.232} & \textbf{28.11\%} &0.262&0.344&1.104& -70.09\% & 0.177& 0.503 & 2.126 & -51.43 &  \textbf{0.762} & \textbf{0.186} & \textbf{1.290} &  \textbf{52.89\%} \\
& \textbf{SM$^4$Depth}                                   & 0.869 & 0.127 & 0.301 & 9.43\% & \underline{0.790} & \textbf{0.134} & \underline{0.673} & \underline{15.06\%} & \underline{0.527} & \textbf{0.233} & \textbf{1.407} & \textbf{18.99\%} & 0.356 & 0.300 & 1.721 & 1.04\% \\
\bottomrule
\toprule
\multirow{2}{*}{\textbf{Categories}} & \multirow{2}{*}{\textbf{Method}} & \multicolumn{4}{c|}{\textbf{nuScenes-val}} 
& \multicolumn{4}{c|}{\textbf{DDAD}} & \multicolumn{4}{c|}{\textbf{ETH3D Outdoor}} 
& \multicolumn{4}{c}{\textbf{DIODE Outdoor}} 
\\ &
& ~$\mathbf{\delta_1 \uparrow}$~ & \textbf{REL~$\downarrow$} & \textbf{RMSE~$\downarrow$} & \textbf{mRI$_\theta\uparrow$}
& ~$\mathbf{\delta_1 \uparrow}$~ & \textbf{REL~$\downarrow$} & \textbf{RMSE~$\downarrow$} & \textbf{mRI$_\theta\uparrow$} 
& ~$\mathbf{\delta_1 \uparrow}$~ & \textbf{REL~$\downarrow$} & \textbf{RMSE~$\downarrow$} & \textbf{mRI$_\theta\uparrow$}
& ~$\mathbf{\delta_1 \uparrow}$~ & \textbf{REL~$\downarrow$} & \textbf{RMSE~$\downarrow$} & \textbf{mRI$_\theta\uparrow$} 
\\ \midrule
\multirow{4}{*}{\textbf{Full-shot}} & BTS \cite{BTS2019} & 0.420 & 0.285 & 9.140 & -9.24\% & 0.802 & 0.146 & 7.611 & -13.07\% & 0.175 & 0.831 & 5.746 & 7.19\% & 0.172 & 0.838 & 10.475 & -34.70\%  \\
& AdaBins \cite{AdaBins2021} & 0.483 & 0.272 & 10.178 & -7.45\% & 0.757 & 0.155 & 8.673 & -22.80\% & 0.110 & 0.889 & 6.480 & -12.65\% & 0.162 & 0.853 & 10.322 & -36.09\%  \\
& NeWCRFs \cite{NeWCRFs2022} & 0.415 & 0.280 & 7.402 & -0.64\% & 0.866 & 0.120 & 6.359 & \underline{2.66\%} & 0.258 & 0.799 & 5.061 & 29.57\% & 0.177 & 0.841 & 9.304 & -29.25\% \\ 
& MIM \cite{MIM2022} & 0.396 & 0.283 & 6.868 & 0.00\% & 0.859 & 0.134 & 6.157 & 0.00\% & 0.159 & 0.889 & 6.048 & 0.00\% & 0.269 & 0.625 & 7.819 & 0.00\% \\
\midrule
\multirow{7}{*}{\makecell{\textbf{Zero-shot}}} 
& ZoeD-K \cite{ZoeDepth2023} & 0.379 & 0.290 & 6.900 & -2.41\% & 0.833 & 0.130 & 7.154 & -5.41\% & 0.303 & 1.012 & 5.853 & 26.65\% & 0.269 & 0.823 & \underline{6.891} & -6.60\%  \\
& DepthAnything-K \cite{depthanything} & 0.579 & 0.223 & \underline{5.844} & 27.44\% & 0.840 & \underline{0.118} & 6.953 & -1.06\% & 0.193 & 0.897 & 6.423 & 4.76\% & \underline{0.309} & 0.836 & 7.599 & -5.35\% \\
& \textbf{SM$^4$Depth-K} & 0.623 & 0.229 & 7.175 & 23.98\%  & 0.841 & 0.160 & 5.677 & -4.56\%  & \textbf{0.452} & \underline{0.294} & \textbf{3.168} & \textbf{99.61\%} & \textbf{0.280} & 0.552 & 8.335 & \underline{3.06\%} \\
\cmidrule(lr){2-18}
& ZoeD-NK \cite{ZoeDepth2023} & 0.371 & 0.299 & 6.988 & -4.57\% & 0.821 & 0.139 & 7.274 & -8.77\%  & 0.337 & 0.752 & 4.758 & 49.56\% & 0.207 & 0.735 & 7.570 & -12.49\% \\
& Metric3D*\cite{Metric3d2023} & \underline{0.868} & \underline{0.143} & 8.506 & \underline{48.27\%} & \underline{0.896} & 0.119 & 7.262 & -0.01\% & 0.324 & 0.724 & 9.830 & 19.93\%  & 0.169 & 0.499 & 9.353 & -12.21\% \\
& UniDepth-ViT-L \cite{piccinelli2024unidepth} & \textbf{0.921} & \textbf{0.088} & \textbf{4.270} & \textbf{79.76\%} & \textbf{0.935} & \textbf{0.103} & \textbf{5.062} & \textbf{16.58\%} & \underline{0.424} & 0.341 & 4.060 & \underline{87.05\%}& \textbf{0.597} & \textbf{0.483} & \textbf{5.631} & \textbf{57.54\%} \\
& \textbf{SM$^4$Depth} & 0.672 & 0.214 & 7.221 & 29.65\% & 0.890 & 0.123 & \underline{5.390} & \underline{8.09\%} & 0.348 & \textbf{0.273} & \underline{3.274} & 78.01\% & 0.190 & \underline{0.487} & 8.435 & -5.05\% \\
\bottomrule 
\end{tabular}}
\caption{Quantitative results on zero-shot datasets. 
mRI$_\theta$ denotes the mean relative improvement compared to MIM across all metrics($\delta_1$, REL, RMSE). 
All methods undergo evaluation consistently within a specific region.
The best results are in bold and the second-best ones are underlined. 
$\times$ indicates poor performance. * means that Metric3D was trained on DDAD.}
\label{tab:zeroshot quantitative res}
\vspace{-10pt}
\end{table*}

\begin{table*}
\centering
\resizebox{\linewidth}{!}{
\begin{tabular}{cll|ccccccc|ccccccc}
\toprule
\multirow{2}{*}{\textbf{Categories}}  &\multirow{2}{*}{\textbf{Method}} & \multirow{2}{*}{\textbf{Backbone}} & \multicolumn{7}{c|}{\textbf{NYUD}} & \multicolumn{7}{c}{\textbf{KITTI}}  \\
& & & ~$\mathbf{\delta_1 \uparrow}$~ & ~$\mathbf{\delta_2 \uparrow}$~ & ~$\mathbf{\delta_3 \uparrow}$~ & \textbf{REL$\downarrow$} & \textbf{RMSE$\downarrow$} & \textbf{log10$\downarrow$} & \textbf{mRI}$_\theta$ & ~$\mathbf{\delta_1 \uparrow}$~ & ~$\mathbf{\delta_2 \uparrow}$~ & ~$\mathbf{\delta_3 \uparrow}$~ & \textbf{REL$\downarrow$} & \textbf{RMSE$\downarrow$} & \textbf{log10$\downarrow$} & \textbf{mRI}$_\theta$\\
\midrule
\multirow{4}{*}{\textbf{Full-shot}} 
& ZoeD-N/K \cite{ZoeDepth2023} & BEiT-L & 0.956 & {0.995} & \underline{0.999} & {0.075} & {0.279} & {0.032} & 0.00\% & \textbf{0.978}& \textbf{0.998} & \underline{0.999} & \textbf{0.049} & \textbf{2.221} & \textbf{0.021} & \textbf{0.00\%} \\ 
& ZoeD-NK \cite{ZoeDepth2023} & BEiT-L & 0.954 & {0.996} & \underline{0.999} & {0.076} & {0.286} & {0.033} & -1.18\%  & {0.971} & 0.994 & 0.996 & \underline{0.053} & \underline{2.415} & 0.024 & -5.43\% \\
& DepthAnything-N/K \cite{depthanything} & ViT-L & \underline{0.983} & \textbf{0.998} & \textbf{1.000} & \underline{0.055} & \underline{0.212} & \underline{0.024} & \underline{13.15\%} & \underline{0.975} & \underline{0.996} & \textbf{1.000} & 0.057 & 2.443 & 0.024 & -6.83\% \\ 
& \textbf{SM$^4$Depth-N/K} & Swin-B & 0.932 & 0.991 & \underline{0.998} & 0.088 & 0.328 & 0.038 & -9.44\%  & {0.971} & \underline{0.996} & \underline{0.999} & 0.054 & 2.477 & \underline{0.023} & -5.36\% \\
\midrule
\multirow{3}{*}{\textbf{Zero-shot}} & Metric3D \cite{Metric3d2023} & CNXT-L & 0.926 & 0.984 & 0.995 & 0.091 & 0.340 & 0.038 & -11.09\% & 0.962 & 0.993 & {0.998} & 0.060 & 2.969 & 0.026 & -13.69\% \\ 
& UniDepth & ViT-L & \textbf{0.984} &\underline{ 0.997} & \textbf{1.000} & \textbf{0.053} & \textbf{0.208} & \textbf{0.023} &\textbf{14.35\%} & \underline{0.975} & \underline{0.996} & \textbf{1.000} & \textbf{0.049} & 2.476 & \textbf{0.021} & \underline{-1.98\%} \\
& \textbf{SM$^4$Depth} & Swin-B & 0.860 & 0.981 & 0.997 & 0.126 & 0.417 & 0.052 & -31.93\%  & 0.928 & 0.985 & 0.996 & 0.087 & 3.272 & 0.038 & -35.42\% \\
\bottomrule
\end{tabular}}
\caption{\textbf{Quantitative result on NYUD and KITTI}. 
All methods undergo evaluation in a consistent region. 
The best results are in bold and the second-best ones are underlined.}
\label{tab:quantitative res}
\vspace{-10pt}
\end{table*}

\begin{figure*}[t!]
\centering
\includegraphics[width=1\linewidth]{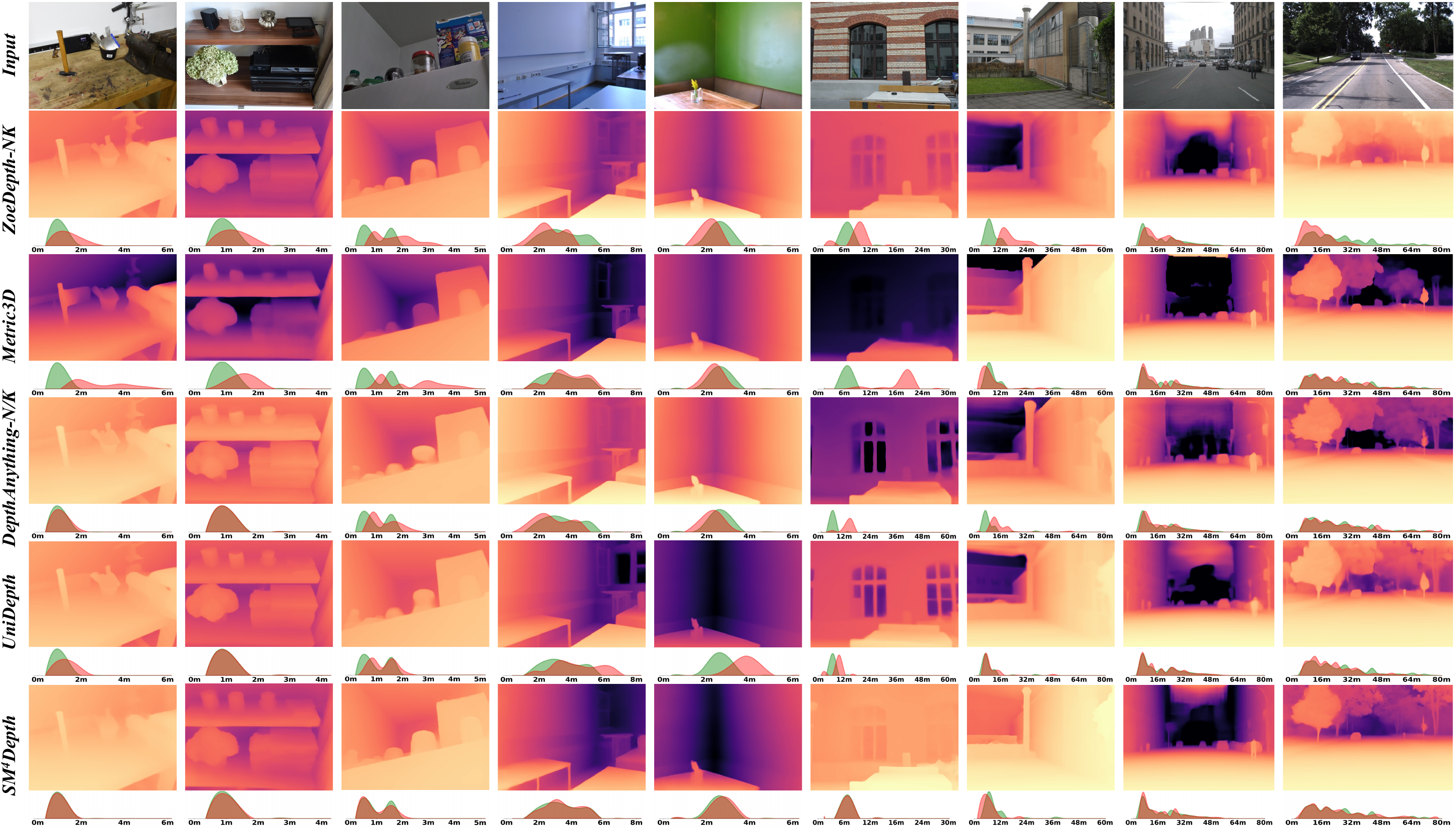}
\caption{
Qualitative comparison with MDE methods on zero-shot datasets.
The depth distribution is under the depth maps with \linegreen{green} for ground truth and \good{red} for prediction.}
\label{fig:qualitative result v3}
\end{figure*}

\subsection{Comparison to the state of the art}

\subsubsection{Quantitative Result}
\label{sec:quantitative result}

We employ two classical MMDE methods, i.e., BTS \cite{BTS2019} and AdaBins \cite{AdaBins2021},
as well as two more advanced MMDE approaches,
i.e., NeWCRFs \cite{NeWCRFs2022} and MIM \cite{mim}, for comparison.
Moreover, we also employ several universal MMDE methods,
i.e., Metric3D \cite{Metric3d2023},
ZoeDepth \cite{ZoeDepth2023},
DepthAnything \cite{depthanything},
and UniDepth \cite{piccinelli2024unidepth},
for comparison
(N indicates NYUD fine-tuning and K for KITTI fine-tuning;
they are also applicable to SM$^4$Depth).

In Table \ref{tab:zeroshot quantitative res},
the upper part shows the zero-shot performance on four indoor datasets,
and the lower part shows that on four outdoor datasets.
SM$^4$Depth not only achieves accuracy consistency across indoor and outdoor but also outperforms most MMDE methods on public datasets and competes with top-performing algorithms like UniDepth and DepthAnything.
Compared to the earliest universal MMDE algorithm, ZoeDepth,
SM$^4$Depth demonstrates superior accuracy across all datasets,
leading in both absolute metrics ($\delta_1$, RMSE) and relative metrics (REL).
This indicates SM$^4$Depth's ability to learn more accurate relative depth from metric depth data.
Compared to Metric3D, SM$^4$Depth performs better on most datasets, (i.e., SUN RGB-D, ETH3D, DIODE, and DDAD) 
and similar on iBims-1,
but is only trained 150K images,
which proves the effectiveness of SM$^4$Depth.
Especially, SM$^4$Depth outperforms Metric3D by +58.08\% and +8.10\% mRI$_\theta$ on ETH3D Outdoor and DDAD. 
In addition, SM$^4$Depth outperforms Metric3D on nuScenes-val by -1.285 of RMSE,
but falls behind on $\delta_1$ and REL,
as Metric3D is trained on much more self-driving datasets,
which endows it an advantage in such scenes.
Notably, SM$^4$Depth outperforms Metric3D by +58.08\% and +8.10\% mRI$_\theta$ on ETH3D Outdoor and DDAD, respectively.
Although not surpassing UniDepth in accuracy overall,
SM$^4$Depth closely approaches or even exceeds it on some datasets (iBims-1, ETH3D Indoor/Outdoor).

Table \ref{tab:quantitative res} displays the results on NYUD and KITTI.
With the zero-shot setting,
our method obtains lower $\delta_1$ and higher RMSE than Metric3D on NYUD and KITTI.
However, after being fine-tuned on NYUD and KITTI,
SM$^4$Depth achieves competitive accuracy with the state-of-the-art methods,
while avoiding a significant degradation in accuracy on zero-shot datasets
(see in Table \ref{tab:zeroshot quantitative res}).

\subsubsection{Qualitative Result}
Fig. \ref{fig:qualitative result v3} visualizes several methods' predictions and depth distributions. 
The $1^\text{st}-3^\text{rd}$ columns show close-up scenes challenging depth range determination.
Previous methods obtain incorrect depth distributions,
while Metric3D tends to push the background farther when the foreground boundary is clearly delineated.
The $4^\text{th}$ and $5^\text{th}$ columns show indoor scenes containing a large area of wall.
Other methods suffer from incorrect depth range,
while SM$^4$Depth recovers the depth distribution accurately.
The $6^\text{th}$ and $7^\text{th}$ columns show two close-up outdoor scenes. 
The predictions of ZoeDepth and DepthAnything exhibit overall shifts, 
while Metric3D fails to distinctly differentiate between the front objects and the wall. 
Due to training on multiple metric depth datasets,
SM$^4$Depth generates a visually reasonable depth distribution
while it does not assign an extreme depth value to sky regions because they are set to 0 during training.
The last two columns show images from self-driving scenes.
Although all methods generate good depth maps,
SM$^4$Depth obtains a more accurate depth distribution and captures richer details than other methods.
Especially in the $9^\text{th}$ column,
where objects are up to 80m away,
our method correctly predicts their farthest depths as well as generating fine tree trunk edges.

\begin{figure}[t]
\centering
\includegraphics[width=1\linewidth]{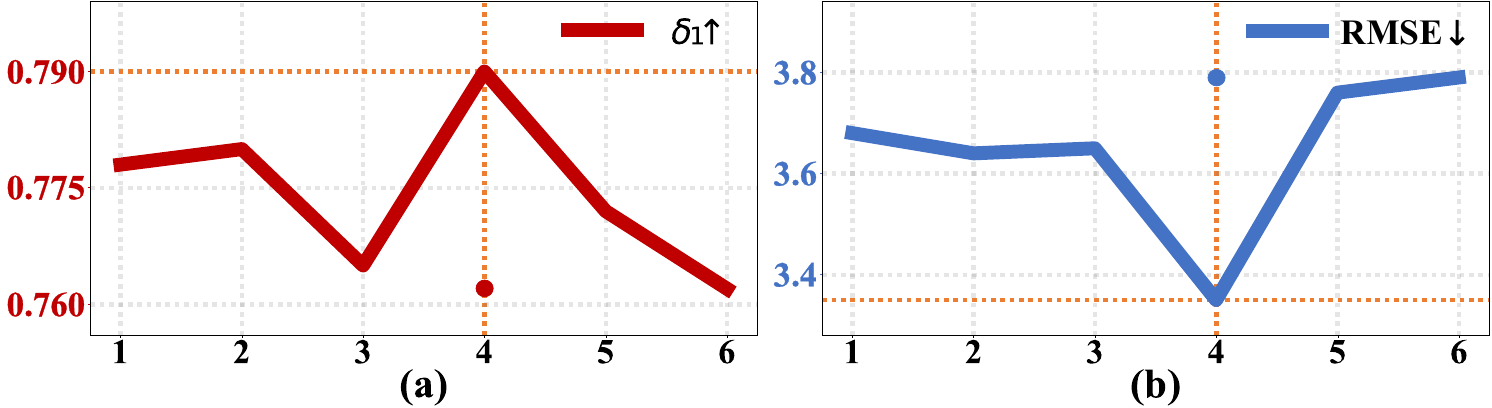}
\caption{
\textbf{Parameter experiment about $K$.}
The performance, as measured by $\delta_1$ and RMSE, 
is optimal when $K$ equals 4.
The dots represent the use of the uniform partition strategy.}
\label{fig:k ablation}
\end{figure}

\begin{table}
\centering
\resizebox{\linewidth}{!}{
\begin{tabular}{c|cc|c|cccc|c}
\toprule
\textbf{\ \ \ V-Bin$^1$} & \textbf{WF-Bin$^{2.1}$} & \textbf{DBE$^{2.2}$} & \textbf{HSC$^3$}  
& \textbf{iBims-1} & \textbf{ETH3D} & \textbf{DIODE} & \textbf{DDAD} 
& \textbf{mRI$_\eta$ $\uparrow$} \\
\midrule
$\surd$  & $\surd$  & $\surd$  & $\surd$  
& \textbf{0.673} & \textbf{2.373} & \textbf{5.605} & \textbf{5.390} & \textbf{10.92\%} \\
$\times$ & $\surd$  & $\surd$  & $\surd$ 
& \underline{0.692} & \underline{2.504} & \underline{6.033} & 5.726 & \ \ \underline{6.03\%} \\
$\times$ & $\times$ & $\surd$  & $\surd$
& 0.701 & 2.692 & 6.111 & \underline{5.486} & \ \ 4.53\% \\
$\times$ & $\times$ & $\times$ & $\surd$
& 0.741 & 2.566 & 6.163 & 5.587 & \ \ 3.67\% \\ 
$\times$ & $\times$ & $\times$ & $\times$
& 0.695 & 2.695 & 6.107 & 6.767 & \ \ 0.00\% \\ \bottomrule
\multicolumn{3}{l}{1: \textit{Depth-Variation based Bin}} &
\multicolumn{5}{l}{2.2: \textit{Domain-aware Bin Estimation}} \\
\multicolumn{3}{l}{2.1: \textit{Weighted Fusion of Bins}} &
\multicolumn{5}{l}{3: \textit{Decoder with Hierarchical Scale Constraints}} \\
\end{tabular}}
\caption{\textbf{RMSE results of the ablation study.}
The best results are in bold, while the second-best ones are underlined.}
\label{tab:module ablation}
\vspace{-14pt}
\end{table}

\subsection{Detail Analysis}
\label{sec:detail analysis}

\subsubsection{Number of depth range domain}
We explore the optimal number of RD, i.e., $K$, and additionally evaluate the uniform partition strategy \cite{DORN_2018_CVPR} 
when using the best $K$.
Fig. \ref{fig:k ablation} shows all variants' performance on the mixing test sets.
As $K$ increases,
RMSE decreases slowly.
RMSE suddenly drops below 3.4 when $K$ = 4 and increases again at 5 or 6.
We argue that this phenomenon occurs because RDs better describe images with different appearance when $K$ = 4 and prevent excessive similarity between RDs due to redundant division.
In addition,
using the uniform partition strategy
leads to a notable decrease in $\delta_1$ and RMSE.

\subsubsection{Ablation study}
We conduct the ablation study by gradually removing our designs
and comparing all variants on the mixing test sets. 
In Table \ref{tab:module ablation},
the baseline (last row) consists of only an encoder-decoder structure and a Pyramid Scene Transformer \cite{DANet2021}.
Observably, the RMSEs increase overall as the proposed modules and innovations are gradually removed.
The depth-variation based bins make the greatest contribution (+4.89\% mRI$_\eta$),
indicating its effectiveness in learning large depth range gaps.
The entire domain-aware bin estimation increases mRI$_\eta$ by 2.36\%,
with the weighted fusion scheme contributing 1.5\% of this.
In addition, the HSC-decoder improves mRI$_\eta$ by 3.67\%.

\subsubsection{Comparing designs for domain-aware bin estimation}
As shown in Table \ref{tab:dbe ablation}, we compare three design choices of our domain-aware bin estimation mentioned in Sec.\ref{sec:dbed} 
on the same four datasets in the ablation study.
Compared to the other settings,
``$K$*Query+$1$*FFN'' achieves the lowest RMSE and highest mRI$_\eta$,
and outperforms other variants by a large margin.
The reason is that the single FFN is trained on multiple RDs and thus learns common knowledge for bin estimation from multiple RDs.

\begin{table}
\centering
\resizebox{\linewidth}{!}{
\begin{tabular}{l|cccc|c}
\toprule
\textbf{Design Choices} & \textbf{iBims-1} & \textbf{ETH3D} & \textbf{DIODE} & \textbf{DDAD} 
& \textbf{mRI$_\eta$ $\uparrow$} \\
\midrule
$1$\ \  * Query\ \ \ + $K$ * FFNs & 0.770 & 2.522 & 5.982 & \underline{6.601} & \ \ 0.00\% \\
$K$ * Queries + $K$ * FFNs & \underline{0.734} & \underline{2.401} & \underline{5.820} & 6.920 
& \ \ \underline{1.84\%}\\
$K$ * Queries + $1$\ \ * FFN & \textbf{0.673} & \textbf{2.373} & \textbf{5.605} & \textbf{5.390} 
& \textbf{10.79\%} \\ \bottomrule
\end{tabular}}
\caption{\textbf{RMSE results of DBE.}
The best results are in bold, while the second-best ones are underlined.}
\label{tab:dbe ablation}
\vspace{-30pt}
\end{table}

\section{Conclusion}
This paper proposes a seamless MMDE algorithm, SM$^4$Depth,
to solve the problems of inconsistent accuracy across diverse scenes and reliance on massive training data.
Firstly, we discuss the inherent issue of the bin-based methods when learning depth range with large gap,
that is the large inconsistency of the same bin in different images.
To address this issue,
we propose the variation-based depth bins that allow the network to effectively learn scenes with different depth ranges.
Next, to reduce the complexity of estimating correct metric bins from a vast solution space,
this paper designs a “divide and conquer” method to determine metric bins from multiple solution sub-spaces,
thereby reducing the network's reliance on massive training data.
Finally, we propose an uncut depth dataset, BUPT Depth, to verify the accuracy consistency across scenes.
Our method obtains outstanding performance with only 150K RGB-D pairs for training and achieves accuracy consistency.
\section{Acknowledgment}
This work was supported by the Innovation Research Group Project of NSFC (61921003), the MUR PNRR project FAIR (PE00000013) funded by the NextGenerationEU and by the PRIN project CREATIVE (Prot.2020ZSL9F9).
\bibliographystyle{ACM-Reference-Format}
\bibliography{main_camera_ready}


\begin{thebibliography}{76}


\ifx \showCODEN    \undefined \def \showCODEN     #1{\unskip}     \fi
\ifx \showDOI      \undefined \def \showDOI       #1{#1}\fi
\ifx \showISBNx    \undefined \def \showISBNx     #1{\unskip}     \fi
\ifx \showISBNxiii \undefined \def \showISBNxiii  #1{\unskip}     \fi
\ifx \showISSN     \undefined \def \showISSN      #1{\unskip}     \fi
\ifx \showLCCN     \undefined \def \showLCCN      #1{\unskip}     \fi
\ifx \shownote     \undefined \def \shownote      #1{#1}          \fi
\ifx \showarticletitle \undefined \def \showarticletitle #1{#1}   \fi
\ifx \showURL      \undefined \def \showURL       {\relax}        \fi
\providecommand\bibfield[2]{#2}
\providecommand\bibinfo[2]{#2}
\providecommand\natexlab[1]{#1}
\providecommand\showeprint[2][]{arXiv:#2}

\bibitem[Anthes et~al\mbox{.}(2016)]%
        {7500674}
\bibfield{author}{\bibinfo{person}{Christoph Anthes},
  \bibinfo{person}{Rubén~Jesús García-Hernández}, \bibinfo{person}{Markus
  Wiedemann}, {and} \bibinfo{person}{Dieter Kranzlmüller}.}
  \bibinfo{year}{2016}\natexlab{}.
\newblock \showarticletitle{State of the art of virtual reality technology}. In
  \bibinfo{booktitle}{\emph{IEEE Aerospace Conference}}.
  \bibinfo{pages}{1--19}.
\newblock


\bibitem[Bauer et~al\mbox{.}(2019)]%
        {uasol}
\bibfield{author}{\bibinfo{person}{Zuria Bauer}, \bibinfo{person}{Francisco
  Gomez-Donoso}, \bibinfo{person}{Edmanuel Cruz}, \bibinfo{person}{Sergio
  Orts-Escolano}, {and} \bibinfo{person}{Miguel Cazorla}.}
  \bibinfo{year}{2019}\natexlab{}.
\newblock \showarticletitle{UASOL, a large-scale high-resolution outdoor stereo
  dataset}.
\newblock \bibinfo{journal}{\emph{Scientific data}} \bibinfo{volume}{6},
  \bibinfo{number}{1} (\bibinfo{year}{2019}), \bibinfo{pages}{162}.
\newblock


\bibitem[Bhat et~al\mbox{.}(2021)]%
        {AdaBins2021}
\bibfield{author}{\bibinfo{person}{Shariq~Farooq Bhat},
  \bibinfo{person}{Ibraheem Alhashim}, {and} \bibinfo{person}{Peter Wonka}.}
  \bibinfo{year}{2021}\natexlab{}.
\newblock \showarticletitle{AdaBins: Depth Estimation Using Adaptive Bins}. In
  \bibinfo{booktitle}{\emph{IEEE Conference on Computer Vision and Pattern
  Recognition (CVPR)}}.
\newblock


\bibitem[Bhat et~al\mbox{.}(2022)]%
        {LocalBins2022}
\bibfield{author}{\bibinfo{person}{Shariq~Farooq Bhat},
  \bibinfo{person}{Ibraheem Alhashim}, {and} \bibinfo{person}{Peter Wonka}.}
  \bibinfo{year}{2022}\natexlab{}.
\newblock \showarticletitle{LocalBins: Improving Depth Estimation by Learning
  Local Distributions}. In \bibinfo{booktitle}{\emph{Computer Vision--ECCV
  2022: 17th European Conference, Tel Aviv, Israel, October 23--27, 2022,
  Proceedings, Part I}}. Springer, \bibinfo{pages}{480--496}.
\newblock


\bibitem[Bhat et~al\mbox{.}(2023)]%
        {ZoeDepth2023}
\bibfield{author}{\bibinfo{person}{Shariq~Farooq Bhat}, \bibinfo{person}{Reiner
  Birkl}, \bibinfo{person}{Diana Wofk}, \bibinfo{person}{Peter Wonka}, {and}
  \bibinfo{person}{Matthias Müller}.} \bibinfo{year}{2023}\natexlab{}.
\newblock \bibinfo{title}{ZoeDepth: Zero-shot Transfer by Combining Relative
  and Metric Depth}.
\newblock
\newblock
\urldef\tempurl%
\url{https://doi.org/10.48550/ARXIV.2302.12288}
\showDOI{\tempurl}


\bibitem[Caesar et~al\mbox{.}(2020)]%
        {Nuscenes2020}
\bibfield{author}{\bibinfo{person}{Holger Caesar}, \bibinfo{person}{Varun
  Bankiti}, \bibinfo{person}{Alex~H Lang}, \bibinfo{person}{Sourabh Vora},
  \bibinfo{person}{Venice~Erin Liong}, \bibinfo{person}{Qiang Xu},
  \bibinfo{person}{Anush Krishnan}, \bibinfo{person}{Yu Pan},
  \bibinfo{person}{Giancarlo Baldan}, {and} \bibinfo{person}{Oscar Beijbom}.}
  \bibinfo{year}{2020}\natexlab{}.
\newblock \showarticletitle{nuscenes: A multimodal dataset for autonomous
  driving}. In \bibinfo{booktitle}{\emph{Proceedings of the IEEE/CVF conference
  on computer vision and pattern recognition}}. \bibinfo{pages}{11621--11631}.
\newblock


\bibitem[Chai et~al\mbox{.}(2023)]%
        {chai2023stablevideo}
\bibfield{author}{\bibinfo{person}{Wenhao Chai}, \bibinfo{person}{Xun Guo},
  \bibinfo{person}{Gaoang Wang}, {and} \bibinfo{person}{Yan Lu}.}
  \bibinfo{year}{2023}\natexlab{}.
\newblock \showarticletitle{Stablevideo: Text-driven consistency-aware
  diffusion video editing}. In \bibinfo{booktitle}{\emph{Proceedings of the
  IEEE/CVF International Conference on Computer Vision}}.
\newblock


\bibitem[Chang et~al\mbox{.}(2022)]%
        {chang2022fast}
\bibfield{author}{\bibinfo{person}{Yicong Chang}, \bibinfo{person}{Feng Xue},
  \bibinfo{person}{Fei Sheng}, \bibinfo{person}{Wenteng Liang}, {and}
  \bibinfo{person}{Anlong Ming}.} \bibinfo{year}{2022}\natexlab{}.
\newblock \showarticletitle{Fast road segmentation via uncertainty-aware
  symmetric network}. In \bibinfo{booktitle}{\emph{International Conference on
  Robotics and Automation (ICRA)}}. \bibinfo{pages}{11124--11130}.
\newblock


\bibitem[Chen et~al\mbox{.}(2019)]%
        {SARPN2019}
\bibfield{author}{\bibinfo{person}{Xiaotian Chen}, \bibinfo{person}{Xuejin
  Chen}, {and} \bibinfo{person}{Zheng-Jun Zha}.}
  \bibinfo{year}{2019}\natexlab{}.
\newblock \showarticletitle{Structure-Aware Residual Pyramid Network for
  Monocular Depth Estimation}. In \bibinfo{booktitle}{\emph{International Joint
  Conference on Artificial Intelligence (IJCAI)}}.
\newblock


\bibitem[Chen et~al\mbox{.}(2022)]%
        {ViT-Adapter2022}
\bibfield{author}{\bibinfo{person}{Zhe Chen}, \bibinfo{person}{Yuchen Duan},
  \bibinfo{person}{Wenhai Wang}, \bibinfo{person}{Junjun He},
  \bibinfo{person}{Tong Lu}, \bibinfo{person}{Jifeng Dai}, {and}
  \bibinfo{person}{Yu Qiao}.} \bibinfo{year}{2022}\natexlab{}.
\newblock \showarticletitle{Vision transformer adapter for dense predictions}.
\newblock \bibinfo{journal}{\emph{arXiv preprint arXiv:2205.08534}}
  (\bibinfo{year}{2022}).
\newblock


\bibitem[Chen et~al\mbox{.}(2021)]%
        {10.1145/3474085.3475287}
\bibfield{author}{\bibinfo{person}{Zhi Chen}, \bibinfo{person}{Xiaoqing Ye},
  \bibinfo{person}{Liang Du}, \bibinfo{person}{Wei Yang},
  \bibinfo{person}{Liusheng Huang}, \bibinfo{person}{Xiao Tan},
  \bibinfo{person}{Zhenbo Shi}, \bibinfo{person}{Fumin Shen}, {and}
  \bibinfo{person}{Errui Ding}.} \bibinfo{year}{2021}\natexlab{}.
\newblock \showarticletitle{AggNet for Self-supervised Monocular Depth
  Estimation: Go An Aggressive Step Further}. In
  \bibinfo{booktitle}{\emph{Proceedings of the 29th ACM International
  Conference on Multimedia}}.
\newblock


\bibitem[Cho et~al\mbox{.}(2021)]%
        {diml}
\bibfield{author}{\bibinfo{person}{Jaehoon Cho}, \bibinfo{person}{Dongbo Min},
  \bibinfo{person}{Youngjung Kim}, {and} \bibinfo{person}{Kwanghoon Sohn}.}
  \bibinfo{year}{2021}\natexlab{}.
\newblock \showarticletitle{{DIML/CVL RGB-D} dataset: 2M RGB-D images of
  natural indoor and outdoor scenes}.
\newblock \bibinfo{journal}{\emph{arXiv preprint arXiv:2110.11590}}
  (\bibinfo{year}{2021}).
\newblock


\bibitem[Chu and Hsu(2018)]%
        {Chu}
\bibfield{author}{\bibinfo{person}{Wei-Ta Chu} {and} \bibinfo{person}{Yu-Ting
  Hsu}.} \bibinfo{year}{2018}\natexlab{}.
\newblock \showarticletitle{Depth-Aware Image Colorization Network}. In
  \bibinfo{booktitle}{\emph{ACMMM Workshop}}.
\newblock


\bibitem[Cordts et~al\mbox{.}(2016)]%
        {cityscapes}
\bibfield{author}{\bibinfo{person}{Marius Cordts}, \bibinfo{person}{Mohamed
  Omran}, \bibinfo{person}{Sebastian Ramos}, \bibinfo{person}{Timo Rehfeld},
  \bibinfo{person}{Markus Enzweiler}, \bibinfo{person}{Rodrigo Benenson},
  \bibinfo{person}{Uwe Franke}, \bibinfo{person}{Stefan Roth}, {and}
  \bibinfo{person}{Bernt Schiele}.} \bibinfo{year}{2016}\natexlab{}.
\newblock \showarticletitle{The Cityscapes Dataset for Semantic Urban Scene
  Understanding}. In \bibinfo{booktitle}{\emph{Proc. of the IEEE Conference on
  Computer Vision and Pattern Recognition (CVPR)}}.
\newblock


\bibitem[Dai et~al\mbox{.}(2017)]%
        {scannet}
\bibfield{author}{\bibinfo{person}{Angela Dai}, \bibinfo{person}{Angel~X.
  Chang}, \bibinfo{person}{Manolis Savva}, \bibinfo{person}{Maciej Halber},
  \bibinfo{person}{Thomas Funkhouser}, {and} \bibinfo{person}{Matthias
  Nie{\ss}ner}.} \bibinfo{year}{2017}\natexlab{}.
\newblock \showarticletitle{ScanNet: Richly-annotated 3D Reconstructions of
  Indoor Scenes}. In \bibinfo{booktitle}{\emph{Proc. Computer Vision and
  Pattern Recognition (CVPR), IEEE}}.
\newblock


\bibitem[Duan et~al\mbox{.}(2018)]%
        {8486539}
\bibfield{author}{\bibinfo{person}{Xiangyue Duan}, \bibinfo{person}{Xinchen
  Ye}, \bibinfo{person}{Yang Li}, {and} \bibinfo{person}{Haojie Li}.}
  \bibinfo{year}{2018}\natexlab{}.
\newblock \showarticletitle{High Quality Depth Estimation from Monocular Images
  Based on Depth Prediction and Enhancement Sub-Networks}. In
  \bibinfo{booktitle}{\emph{2018 IEEE International Conference on Multimedia
  and Expo (ICME)}}.
\newblock


\bibitem[Dukor et~al\mbox{.}(2022)]%
        {10.1145/3532719.3543235}
\bibfield{author}{\bibinfo{person}{Obumneme~Stanley Dukor},
  \bibinfo{person}{S.~Mahdi H.~Miangoleh}, \bibinfo{person}{Mahesh~Kumar
  Krishna~Reddy}, \bibinfo{person}{Long Mai}, {and}
  \bibinfo{person}{Ya\u{g}\i{}z Aksoy}.} \bibinfo{year}{2022}\natexlab{}.
\newblock \showarticletitle{Interactive Editing of Monocular Depth}. In
  \bibinfo{booktitle}{\emph{ACM SIGGRAPH 2022 Posters}}.
\newblock


\bibitem[Fan et~al\mbox{.}(2017)]%
        {ChamferLoss2017}
\bibfield{author}{\bibinfo{person}{Haoqiang Fan}, \bibinfo{person}{Hao Su},
  {and} \bibinfo{person}{Leonidas Guibas}.} \bibinfo{year}{2017}\natexlab{}.
\newblock \showarticletitle{A Point Set Generation Network for 3D Object
  Reconstruction from a Single Image}. In \bibinfo{booktitle}{\emph{2017 IEEE
  Conference on Computer Vision and Pattern Recognition (CVPR)}}.
\newblock
\urldef\tempurl%
\url{https://doi.org/10.1109/cvpr.2017.264}
\showDOI{\tempurl}


\bibitem[Fu et~al\mbox{.}(2018)]%
        {DORN_2018_CVPR}
\bibfield{author}{\bibinfo{person}{Huan Fu}, \bibinfo{person}{Mingming Gong},
  \bibinfo{person}{Chaohui Wang}, \bibinfo{person}{Kayhan Batmanghelich}, {and}
  \bibinfo{person}{Dacheng Tao}.} \bibinfo{year}{2018}\natexlab{}.
\newblock \showarticletitle{Deep Ordinal Regression Network for Monocular Depth
  Estimation}. In \bibinfo{booktitle}{\emph{IEEE Conference on Computer Vision
  and Pattern Recognition (CVPR)}}.
\newblock


\bibitem[Guizilini et~al\mbox{.}(2020)]%
        {ddad}
\bibfield{author}{\bibinfo{person}{Vitor Guizilini}, \bibinfo{person}{Rares
  Ambrus}, \bibinfo{person}{Sudeep Pillai}, \bibinfo{person}{Allan Raventos},
  {and} \bibinfo{person}{Adrien Gaidon}.} \bibinfo{year}{2020}\natexlab{}.
\newblock \showarticletitle{3D Packing for Self-Supervised Monocular Depth
  Estimation}. In \bibinfo{booktitle}{\emph{IEEE Conference on Computer Vision
  and Pattern Recognition (CVPR)}}.
\newblock


\bibitem[Huang et~al\mbox{.}(2018)]%
        {apolloscape}
\bibfield{author}{\bibinfo{person}{Xinyu Huang}, \bibinfo{person}{Xinjing
  Cheng}, \bibinfo{person}{Qichuan Geng}, \bibinfo{person}{Binbin Cao},
  \bibinfo{person}{Dingfu Zhou}, \bibinfo{person}{Peng Wang},
  \bibinfo{person}{Yuanqing Lin}, {and} \bibinfo{person}{Ruigang Yang}.}
  \bibinfo{year}{2018}\natexlab{}.
\newblock \showarticletitle{The apolloscape dataset for autonomous driving}. In
  \bibinfo{booktitle}{\emph{Proceedings of the IEEE conference on computer
  vision and pattern recognition workshops}}. \bibinfo{pages}{954--960}.
\newblock


\bibitem[Ioannou and Maddock(2022)]%
        {ioannou2022depth}
\bibfield{author}{\bibinfo{person}{Eleftherios Ioannou} {and}
  \bibinfo{person}{Steve Maddock}.} \bibinfo{year}{2022}\natexlab{}.
\newblock \showarticletitle{Depth-aware neural style transfer using instance
  normalization}. In \bibinfo{booktitle}{\emph{Computer Graphics \& Visual
  Computing (CGVC) 2022}}. Eurographics Digital Library.
\newblock


\bibitem[Koch et~al\mbox{.}(2018)]%
        {ibims}
\bibfield{author}{\bibinfo{person}{Tobias Koch}, \bibinfo{person}{Lukas
  Liebel}, \bibinfo{person}{Friedrich Fraundorfer}, {and}
  \bibinfo{person}{Marco Korner}.} \bibinfo{year}{2018}\natexlab{}.
\newblock \showarticletitle{Evaluation of cnn-based single-image depth
  estimation methods}. In \bibinfo{booktitle}{\emph{Proceedings of the European
  Conference on Computer Vision (ECCV) Workshops}}. \bibinfo{pages}{0--0}.
\newblock


\bibitem[Laina et~al\mbox{.}(2016)]%
        {Laina2016}
\bibfield{author}{\bibinfo{person}{Iro Laina}, \bibinfo{person}{Christian
  Rupprecht}, \bibinfo{person}{Vasileios Belagiannis},
  \bibinfo{person}{Federico Tombari}, {and} \bibinfo{person}{Nassir Navab}.}
  \bibinfo{year}{2016}\natexlab{}.
\newblock \showarticletitle{Deeper Depth Prediction with Fully Convolutional
  Residual Networks}. In \bibinfo{booktitle}{\emph{International Conference on
  3D Vision (3DV)}}.
\newblock


\bibitem[Lee et~al\mbox{.}(2019)]%
        {BTS2019}
\bibfield{author}{\bibinfo{person}{Jin~Han Lee}, \bibinfo{person}{Myung-Kyu
  Han}, \bibinfo{person}{Dong~Wook Ko}, {and} \bibinfo{person}{Il~Hong Suh}.}
  \bibinfo{year}{2019}\natexlab{}.
\newblock \showarticletitle{From big to small: Multi-scale local planar
  guidance for monocular depth estimation}.
\newblock \bibinfo{journal}{\emph{arXiv preprint arXiv:1907.10326}}
  (\bibinfo{year}{2019}).
\newblock


\bibitem[Li et~al\mbox{.}(2022b)]%
        {CRESTEREO2022}
\bibfield{author}{\bibinfo{person}{Jiankun Li}, \bibinfo{person}{Peisen Wang},
  \bibinfo{person}{Pengfei Xiong}, \bibinfo{person}{Tao Cai},
  \bibinfo{person}{Ziwei Yan}, \bibinfo{person}{Lei Yang},
  \bibinfo{person}{Jiangyu Liu}, \bibinfo{person}{Haoqiang Fan}, {and}
  \bibinfo{person}{Shuaicheng Liu}.} \bibinfo{year}{2022}\natexlab{b}.
\newblock \showarticletitle{Practical stereo matching via cascaded recurrent
  network with adaptive correlation}. In \bibinfo{booktitle}{\emph{Proceedings
  of the IEEE/CVF Conference on Computer Vision and Pattern Recognition}}.
  \bibinfo{pages}{16263--16272}.
\newblock


\bibitem[Li et~al\mbox{.}(2020)]%
        {10.1145/3394171.3413706}
\bibfield{author}{\bibinfo{person}{Rui Li}, \bibinfo{person}{Xiantuo He},
  \bibinfo{person}{Yu Zhu}, \bibinfo{person}{Xianjun Li},
  \bibinfo{person}{Jinqiu Sun}, {and} \bibinfo{person}{Yanning Zhang}.}
  \bibinfo{year}{2020}\natexlab{}.
\newblock \showarticletitle{Enhancing Self-supervised Monocular Depth
  Estimation via Incorporating Robust Constraints}. In
  \bibinfo{booktitle}{\emph{Proceedings of the 28th ACM International
  Conference on Multimedia}}.
\newblock


\bibitem[Li et~al\mbox{.}(2023)]%
        {9852314}
\bibfield{author}{\bibinfo{person}{Rui Li}, \bibinfo{person}{Danna Xue},
  \bibinfo{person}{Yu Zhu}, \bibinfo{person}{Hao Wu}, \bibinfo{person}{Jinqiu
  Sun}, {and} \bibinfo{person}{Yanning Zhang}.}
  \bibinfo{year}{2023}\natexlab{}.
\newblock \showarticletitle{Self-Supervised Monocular Depth Estimation With
  Frequency-Based Recurrent Refinement}.
\newblock \bibinfo{journal}{\emph{IEEE Transactions on Multimedia}}
  \bibinfo{volume}{25} (\bibinfo{year}{2023}), \bibinfo{pages}{5626--5637}.
\newblock


\bibitem[Li and Snavely(2018)]%
        {megadepth2018}
\bibfield{author}{\bibinfo{person}{Zhengqi Li} {and} \bibinfo{person}{Noah
  Snavely}.} \bibinfo{year}{2018}\natexlab{}.
\newblock \showarticletitle{Megadepth: Learning single-view depth prediction
  from internet photos}. In \bibinfo{booktitle}{\emph{Proceedings of the IEEE
  conference on computer vision and pattern recognition}}.
  \bibinfo{pages}{2041--2050}.
\newblock


\bibitem[Li et~al\mbox{.}(2022a)]%
        {BinsFormer2022}
\bibfield{author}{\bibinfo{person}{Zhenyu Li}, \bibinfo{person}{Xuyang Wang},
  \bibinfo{person}{Xianming Liu}, {and} \bibinfo{person}{Junjun Jiang}.}
  \bibinfo{year}{2022}\natexlab{a}.
\newblock \showarticletitle{Binsformer: Revisiting adaptive bins for monocular
  depth estimation}.
\newblock \bibinfo{journal}{\emph{arXiv preprint arXiv:2204.00987}}
  (\bibinfo{year}{2022}).
\newblock


\bibitem[Liang et~al\mbox{.}(2023)]%
        {liang2023unknown}
\bibfield{author}{\bibinfo{person}{Wenteng Liang}, \bibinfo{person}{Feng Xue},
  \bibinfo{person}{Yihao Liu}, \bibinfo{person}{Guofeng Zhong}, {and}
  \bibinfo{person}{Anlong Ming}.} \bibinfo{year}{2023}\natexlab{}.
\newblock \showarticletitle{Unknown Sniffer for Object Detection: Don't Turn a
  Blind Eye to Unknown Objects}. In \bibinfo{booktitle}{\emph{IEEE/CVF
  Conference on Computer Vision and Pattern Recognition}}.
\newblock


\bibitem[Liu et~al\mbox{.}(2017)]%
        {liu2017depth}
\bibfield{author}{\bibinfo{person}{Xiao-Chang Liu}, \bibinfo{person}{Ming-Ming
  Cheng}, \bibinfo{person}{Yu-Kun Lai}, {and} \bibinfo{person}{Paul~L Rosin}.}
  \bibinfo{year}{2017}\natexlab{}.
\newblock \showarticletitle{Depth-aware neural style transfer}. In
  \bibinfo{booktitle}{\emph{Proceedings of the symposium on non-photorealistic
  animation and rendering}}.
\newblock


\bibitem[Lu et~al\mbox{.}(2019b)]%
        {RuiLu-ICCV2019-OFNet}
\bibfield{author}{\bibinfo{person}{Rui Lu}, \bibinfo{person}{Feng Xue},
  \bibinfo{person}{Menghan Zhou}, \bibinfo{person}{Anlong Ming}, {and}
  \bibinfo{person}{Yu Zhou}.} \bibinfo{year}{2019}\natexlab{b}.
\newblock \showarticletitle{Occlusion-Shared and Feature-Separated Network for
  Occlusion Relationship Reasoning}. In \bibinfo{booktitle}{\emph{IEEE/CVF
  International Conference on Computer Vision (ICCV)}}.
\newblock


\bibitem[Lu et~al\mbox{.}(2019a)]%
        {Shufang}
\bibfield{author}{\bibinfo{person}{Shufang Lu}, \bibinfo{person}{Wei Jiang},
  \bibinfo{person}{Xuefeng Ding}, \bibinfo{person}{Craig~S. Kaplan},
  \bibinfo{person}{Xiaogang Jin}, \bibinfo{person}{Fei Gao}, {and}
  \bibinfo{person}{Jiazhou Chen}.} \bibinfo{year}{2019}\natexlab{a}.
\newblock \showarticletitle{Depth-aware image vectorization and editing}.
\newblock \bibinfo{journal}{\emph{The Visual Computer}} \bibinfo{volume}{35},
  \bibinfo{number}{6} (\bibinfo{year}{2019}), \bibinfo{pages}{1027–1039}.
\newblock


\bibitem[Nathan~Silberman and Fergus(2012)]%
        {nyud}
\bibfield{author}{\bibinfo{person}{Pushmeet~Kohli Nathan~Silberman,
  Derek~Hoiem} {and} \bibinfo{person}{Rob Fergus}.}
  \bibinfo{year}{2012}\natexlab{}.
\newblock \showarticletitle{Indoor Segmentation and Support Inference from RGBD
  Images}. In \bibinfo{booktitle}{\emph{ECCV}}.
\newblock


\bibitem[Nie et~al\mbox{.}(2017)]%
        {8237635}
\bibfield{author}{\bibinfo{person}{Bruce~Xiaohan Nie}, \bibinfo{person}{Ping
  Wei}, {and} \bibinfo{person}{Song-Chun Zhu}.}
  \bibinfo{year}{2017}\natexlab{}.
\newblock \showarticletitle{Monocular 3D Human Pose Estimation by Predicting
  Depth on Joints}. In \bibinfo{booktitle}{\emph{IEEE International Conference
  on Computer Vision (ICCV)}}.
\newblock


\bibitem[Piccinelli et~al\mbox{.}(2024)]%
        {piccinelli2024unidepth}
\bibfield{author}{\bibinfo{person}{Luigi Piccinelli}, \bibinfo{person}{Yung-Hsu
  Yang}, \bibinfo{person}{Christos Sakaridis}, \bibinfo{person}{Mattia Segu},
  \bibinfo{person}{Siyuan Li}, \bibinfo{person}{Luc Van~Gool}, {and}
  \bibinfo{person}{Fisher Yu}.} \bibinfo{year}{2024}\natexlab{}.
\newblock \showarticletitle{UniDepth: Universal Monocular Metric Depth
  Estimation}. In \bibinfo{booktitle}{\emph{IEEE Conference on Computer Vision
  and Pattern Recognition (CVPR)}}.
\newblock


\bibitem[Ranftl et~al\mbox{.}(2021a)]%
        {ranftl2021vision}
\bibfield{author}{\bibinfo{person}{Ren{\'e} Ranftl}, \bibinfo{person}{Alexey
  Bochkovskiy}, {and} \bibinfo{person}{Vladlen Koltun}.}
  \bibinfo{year}{2021}\natexlab{a}.
\newblock \showarticletitle{Vision transformers for dense prediction}. In
  \bibinfo{booktitle}{\emph{Proceedings of the IEEE/CVF International
  Conference on Computer Vision}}. \bibinfo{pages}{12179--12188}.
\newblock


\bibitem[Ranftl et~al\mbox{.}(2021b)]%
        {DPT2021}
\bibfield{author}{\bibinfo{person}{Ren{\'e} Ranftl}, \bibinfo{person}{Alexey
  Bochkovskiy}, {and} \bibinfo{person}{Vladlen Koltun}.}
  \bibinfo{year}{2021}\natexlab{b}.
\newblock \showarticletitle{Vision transformers for dense prediction}. In
  \bibinfo{booktitle}{\emph{Proceedings of the IEEE/CVF International
  Conference on Computer Vision}}. \bibinfo{pages}{12179--12188}.
\newblock


\bibitem[Ranftl et~al\mbox{.}(2020)]%
        {MiDaS2020}
\bibfield{author}{\bibinfo{person}{Ren{\'e} Ranftl}, \bibinfo{person}{Katrin
  Lasinger}, \bibinfo{person}{David Hafner}, \bibinfo{person}{Konrad
  Schindler}, {and} \bibinfo{person}{Vladlen Koltun}.}
  \bibinfo{year}{2020}\natexlab{}.
\newblock \showarticletitle{Towards robust monocular depth estimation: Mixing
  datasets for zero-shot cross-dataset transfer}.
\newblock \bibinfo{journal}{\emph{IEEE transactions on pattern analysis and
  machine intelligence}} \bibinfo{volume}{44}, \bibinfo{number}{3}
  (\bibinfo{year}{2020}), \bibinfo{pages}{1623--1637}.
\newblock


\bibitem[Roberts et~al\mbox{.}(2021)]%
        {Hypersim}
\bibfield{author}{\bibinfo{person}{Mike Roberts}, \bibinfo{person}{Jason
  Ramapuram}, \bibinfo{person}{Anurag Ranjan}, \bibinfo{person}{Atulit Kumar},
  \bibinfo{person}{Miguel~Angel Bautista}, \bibinfo{person}{Nathan Paczan},
  \bibinfo{person}{Russ Webb}, {and} \bibinfo{person}{Joshua~M. Susskind}.}
  \bibinfo{year}{2021}\natexlab{}.
\newblock \showarticletitle{{Hypersim}: {A} Photorealistic Synthetic Dataset
  for Holistic Indoor Scene Understanding}. In
  \bibinfo{booktitle}{\emph{International Conference on Computer Vision (ICCV)
  2021}}.
\newblock


\bibitem[Sch\"ops et~al\mbox{.}(2017a)]%
        {schoeps2017cvpr}
\bibfield{author}{\bibinfo{person}{Thomas Sch\"ops},
  \bibinfo{person}{Johannes~L. Sch\"onberger}, \bibinfo{person}{Silvano
  Galliani}, \bibinfo{person}{Torsten Sattler}, \bibinfo{person}{Konrad
  Schindler}, \bibinfo{person}{Marc Pollefeys}, {and} \bibinfo{person}{Andreas
  Geiger}.} \bibinfo{year}{2017}\natexlab{a}.
\newblock \showarticletitle{A Multi-View Stereo Benchmark with High-Resolution
  Images and Multi-Camera Videos}. In \bibinfo{booktitle}{\emph{IEEE Conference
  on Computer Vision and Pattern Recognition (CVPR)}}.
\newblock


\bibitem[Sch\"ops et~al\mbox{.}(2017b)]%
        {eth3d}
\bibfield{author}{\bibinfo{person}{Thomas Sch\"ops},
  \bibinfo{person}{Johannes~L. Sch\"onberger}, \bibinfo{person}{Silvano
  Galliani}, \bibinfo{person}{Torsten Sattler}, \bibinfo{person}{Konrad
  Schindler}, \bibinfo{person}{Marc Pollefeys}, {and} \bibinfo{person}{Andreas
  Geiger}.} \bibinfo{year}{2017}\natexlab{b}.
\newblock \showarticletitle{A Multi-View Stereo Benchmark with High-Resolution
  Images and Multi-Camera Videos}. In \bibinfo{booktitle}{\emph{Conference on
  Computer Vision and Pattern Recognition (CVPR)}}.
\newblock


\bibitem[Shen et~al\mbox{.}(2021)]%
        {10.1145/3474085.3475639}
\bibfield{author}{\bibinfo{person}{Guibao Shen}, \bibinfo{person}{Yingkui
  Zhang}, \bibinfo{person}{Jialu Li}, \bibinfo{person}{Mingqiang Wei},
  \bibinfo{person}{Qiong Wang}, \bibinfo{person}{Guangyong Chen}, {and}
  \bibinfo{person}{Pheng-Ann Heng}.} \bibinfo{year}{2021}\natexlab{}.
\newblock \showarticletitle{Learning Regularizer for Monocular Depth Estimation
  with Adversarial Guidance}. In \bibinfo{booktitle}{\emph{Proceedings of the
  29th ACM International Conference on Multimedia (ACMMM)}}.
\newblock


\bibitem[Shen et~al\mbox{.}(2023)]%
        {10.1145/3581783.3612033}
\bibfield{author}{\bibinfo{person}{Liao Shen}, \bibinfo{person}{Xingyi Li},
  \bibinfo{person}{Huiqiang Sun}, \bibinfo{person}{Juewen Peng},
  \bibinfo{person}{Ke Xian}, \bibinfo{person}{Zhiguo Cao}, {and}
  \bibinfo{person}{Guosheng Lin}.} \bibinfo{year}{2023}\natexlab{}.
\newblock \showarticletitle{Make-It-4D: Synthesizing a Consistent Long-Term
  Dynamic Scene Video from a Single Image}. In
  \bibinfo{booktitle}{\emph{Proceedings of the 31st ACM International
  Conference on Multimedia}}.
\newblock


\bibitem[Sheng et~al\mbox{.}(2022)]%
        {DANet2021}
\bibfield{author}{\bibinfo{person}{Fei Sheng}, \bibinfo{person}{Feng Xue},
  \bibinfo{person}{Yicong Chang}, \bibinfo{person}{Wenteng Liang}, {and}
  \bibinfo{person}{Anlong Ming}.} \bibinfo{year}{2022}\natexlab{}.
\newblock \showarticletitle{Monocular Depth Distribution Alignment with Low
  Computation}. In \bibinfo{booktitle}{\emph{International Conference on
  Robotics and Automation (ICRA)}}.
\newblock


\bibitem[Song et~al\mbox{.}(2023)]%
        {10.1145/3581783.3611800}
\bibfield{author}{\bibinfo{person}{Liangchen Song}, \bibinfo{person}{Liangliang
  Cao}, \bibinfo{person}{Hongyu Xu}, \bibinfo{person}{Kai Kang},
  \bibinfo{person}{Feng Tang}, \bibinfo{person}{Junsong Yuan}, {and}
  \bibinfo{person}{Zhao Yang}.} \bibinfo{year}{2023}\natexlab{}.
\newblock \showarticletitle{RoomDreamer: Text-Driven 3D Indoor Scene Synthesis
  with Coherent Geometry and Texture}. In \bibinfo{booktitle}{\emph{Proceedings
  of the 31st ACM International Conference on Multimedia}}.
\newblock


\bibitem[Song et~al\mbox{.}(2015)]%
        {sunrgbd}
\bibfield{author}{\bibinfo{person}{Shuran Song}, \bibinfo{person}{Samuel~P
  Lichtenberg}, {and} \bibinfo{person}{Jianxiong Xiao}.}
  \bibinfo{year}{2015}\natexlab{}.
\newblock \showarticletitle{{SUN RGB-D}: A rgb-d scene understanding benchmark
  suite}. In \bibinfo{booktitle}{\emph{Proceedings of the IEEE conference on
  computer vision and pattern recognition}}. \bibinfo{pages}{567--576}.
\newblock


\bibitem[Su et~al\mbox{.}(2019)]%
        {10.1145/3343031.3350930}
\bibfield{author}{\bibinfo{person}{Wen Su}, \bibinfo{person}{Haifeng Zhang},
  \bibinfo{person}{Jia Li}, \bibinfo{person}{Wenzhen Yang}, {and}
  \bibinfo{person}{Zengfu Wang}.} \bibinfo{year}{2019}\natexlab{}.
\newblock \showarticletitle{Monocular Depth Estimation as Regression of
  Classification using Piled Residual Networks}. In
  \bibinfo{booktitle}{\emph{Proceedings of the 27th ACM International
  Conference on Multimedia}}.
\newblock


\bibitem[Sun et~al\mbox{.}(2022)]%
        {sun2022putting}
\bibfield{author}{\bibinfo{person}{Yu Sun}, \bibinfo{person}{Wu Liu},
  \bibinfo{person}{Qian Bao}, \bibinfo{person}{Yili Fu}, \bibinfo{person}{Tao
  Mei}, {and} \bibinfo{person}{Michael~J Black}.}
  \bibinfo{year}{2022}\natexlab{}.
\newblock \showarticletitle{Putting people in their place: Monocular regression
  of 3d people in depth}. In \bibinfo{booktitle}{\emph{Proceedings of the
  IEEE/CVF Conference on Computer Vision and Pattern Recognition}}.
\newblock


\bibitem[Tang et~al\mbox{.}(2021)]%
        {10.1145/3474085.3475373}
\bibfield{author}{\bibinfo{person}{Qi Tang}, \bibinfo{person}{Runmin Cong},
  \bibinfo{person}{Ronghui Sheng}, \bibinfo{person}{Lingzhi He},
  \bibinfo{person}{Dan Zhang}, \bibinfo{person}{Yao Zhao}, {and}
  \bibinfo{person}{Sam Kwong}.} \bibinfo{year}{2021}\natexlab{}.
\newblock \showarticletitle{BridgeNet: A Joint Learning Network of Depth Map
  Super-Resolution and Monocular Depth Estimation}. In
  \bibinfo{booktitle}{\emph{Proceedings of the 29th ACM International
  Conference on Multimedia}}.
\newblock


\bibitem[Tateno et~al\mbox{.}(2017)]%
        {cnnslam}
\bibfield{author}{\bibinfo{person}{K. Tateno}, \bibinfo{person}{F. Tombari},
  \bibinfo{person}{I. Laina}, {and} \bibinfo{person}{N. Navab}.}
  \bibinfo{year}{2017}\natexlab{}.
\newblock \showarticletitle{{CNN-SLAM}: Real-time dense monocular SLAM with
  learned depth prediction}. In \bibinfo{booktitle}{\emph{IEEE Conference on
  Computer Vision and Pattern Recognition (CVPR)}}.
\newblock


\bibitem[Uhrig et~al\mbox{.}(2017)]%
        {kitti}
\bibfield{author}{\bibinfo{person}{Jonas Uhrig}, \bibinfo{person}{Nick
  Schneider}, \bibinfo{person}{Lukas Schneider}, \bibinfo{person}{Uwe Franke},
  \bibinfo{person}{Thomas Brox}, {and} \bibinfo{person}{Andreas Geiger}.}
  \bibinfo{year}{2017}\natexlab{}.
\newblock \showarticletitle{Sparsity Invariant CNNs}. In
  \bibinfo{booktitle}{\emph{International Conference on 3D Vision (3DV)}}.
\newblock


\bibitem[Vasiljevic et~al\mbox{.}(2019)]%
        {diode}
\bibfield{author}{\bibinfo{person}{Igor Vasiljevic}, \bibinfo{person}{Nick
  Kolkin}, \bibinfo{person}{Shanyi Zhang}, \bibinfo{person}{Ruotian Luo},
  \bibinfo{person}{Haochen Wang}, \bibinfo{person}{Falcon~Z. Dai},
  \bibinfo{person}{Andrea~F. Daniele}, \bibinfo{person}{Mohammadreza
  Mostajabi}, \bibinfo{person}{Steven Basart}, \bibinfo{person}{Matthew~R.
  Walter}, {and} \bibinfo{person}{Gregory Shakhnarovich}.}
  \bibinfo{year}{2019}\natexlab{}.
\newblock \showarticletitle{{DIODE}: {A} {D}ense {I}ndoor and {O}utdoor {DE}pth
  {D}ataset}.
\newblock \bibinfo{journal}{\emph{CoRR}}  \bibinfo{volume}{abs/1908.00463}
  (\bibinfo{year}{2019}).
\newblock
\urldef\tempurl%
\url{http://arxiv.org/abs/1908.00463}
\showURL{%
\tempurl}


\bibitem[Wang et~al\mbox{.}(2023)]%
        {10.1145/3581783.3612306}
\bibfield{author}{\bibinfo{person}{Chen Wang}, \bibinfo{person}{Jiadai Sun},
  \bibinfo{person}{Lina Liu}, \bibinfo{person}{Chenming Wu},
  \bibinfo{person}{Zhelun Shen}, \bibinfo{person}{Dayan Wu},
  \bibinfo{person}{Yuchao Dai}, {and} \bibinfo{person}{Liangjun Zhang}.}
  \bibinfo{year}{2023}\natexlab{}.
\newblock \showarticletitle{Digging into Depth Priors for Outdoor Neural
  Radiance Fields}. In \bibinfo{booktitle}{\emph{Proceedings of the 31st ACM
  International Conference on Multimedia}}.
\newblock


\bibitem[Wang and Shen(2018)]%
        {8490975}
\bibfield{author}{\bibinfo{person}{Kaixuan Wang} {and} \bibinfo{person}{Shaojie
  Shen}.} \bibinfo{year}{2018}\natexlab{}.
\newblock \showarticletitle{MVDepthNet: Real-Time Multiview Depth Estimation
  Neural Network}. In \bibinfo{booktitle}{\emph{International Conference on 3D
  Vision (3DV)}}.
\newblock


\bibitem[Wu et~al\mbox{.}(2023a)]%
        {9669049}
\bibfield{author}{\bibinfo{person}{Jipeng Wu}, \bibinfo{person}{Rongrong Ji},
  \bibinfo{person}{Qiang Wang}, \bibinfo{person}{Shengchuan Zhang},
  \bibinfo{person}{Xiaoshuai Sun}, \bibinfo{person}{Yan Wang},
  \bibinfo{person}{Mingliang Xu}, {and} \bibinfo{person}{Feiyue Huang}.}
  \bibinfo{year}{2023}\natexlab{a}.
\newblock \showarticletitle{Fast Monocular Depth Estimation via Side Prediction
  Aggregation with Continuous Spatial Refinement}.
\newblock \bibinfo{journal}{\emph{IEEE Transactions on Multimedia}}
  \bibinfo{volume}{25} (\bibinfo{year}{2023}), \bibinfo{pages}{1204--1216}.
\newblock


\bibitem[Wu et~al\mbox{.}(2023b)]%
        {10.1145/3581783.3611751}
\bibfield{author}{\bibinfo{person}{Zizhang Wu}, \bibinfo{person}{Zhuozheng Li},
  \bibinfo{person}{Zhi-Gang Fan}, \bibinfo{person}{Yunzhe Wu},
  \bibinfo{person}{Jian Pu}, {and} \bibinfo{person}{Xianzhi Li}.}
  \bibinfo{year}{2023}\natexlab{b}.
\newblock \showarticletitle{V2Depth: Monocular Depth Estimation via
  Feature-Level Virtual-View Simulation and Refinement}. In
  \bibinfo{booktitle}{\emph{Proceedings of the 31st ACM International
  Conference on Multimedia}}.
\newblock


\bibitem[Xie et~al\mbox{.}(2022)]%
        {MIM2022}
\bibfield{author}{\bibinfo{person}{Zhenda Xie}, \bibinfo{person}{Zigang Geng},
  \bibinfo{person}{Jingcheng Hu}, \bibinfo{person}{Zheng Zhang},
  \bibinfo{person}{Han Hu}, {and} \bibinfo{person}{Yue Cao}.}
  \bibinfo{year}{2022}\natexlab{}.
\newblock \showarticletitle{Revealing the dark secrets of masked image
  modeling}.
\newblock \bibinfo{journal}{\emph{arXiv preprint arXiv:2205.13543}}
  (\bibinfo{year}{2022}).
\newblock


\bibitem[Xue et~al\mbox{.}(2021)]%
        {depthpr2021}
\bibfield{author}{\bibinfo{person}{Feng Xue}, \bibinfo{person}{Junfeng Cao},
  \bibinfo{person}{Yu Zhou}, \bibinfo{person}{Fei Sheng},
  \bibinfo{person}{Yankai Wang}, {and} \bibinfo{person}{Anlong Ming}.}
  \bibinfo{year}{2021}\natexlab{}.
\newblock \showarticletitle{Boundary-induced and scene-aggregated network for
  monocular depth prediction}.
\newblock \bibinfo{journal}{\emph{Pattern Recognition (PR)}}
  \bibinfo{volume}{115} (\bibinfo{year}{2021}), \bibinfo{pages}{107901}.
\newblock


\bibitem[Xue et~al\mbox{.}(2023)]%
        {xue_indoor_2023}
\bibfield{author}{\bibinfo{person}{Feng Xue}, \bibinfo{person}{Yicong Chang},
  \bibinfo{person}{Tianxi Wang}, \bibinfo{person}{Yu Zhou}, {and}
  \bibinfo{person}{Anlong Ming}.} \bibinfo{year}{2023}\natexlab{}.
\newblock \showarticletitle{Indoor {Obstacle} {Discovery} on {Reflective}
  {Ground} via {Monocular} {Camera}}.
\newblock \bibinfo{journal}{\emph{International Journal of Computer Vision}}
  (\bibinfo{date}{oct} \bibinfo{year}{2023}).
\newblock


\bibitem[{Xue} et~al\mbox{.}(2019)]%
        {ICRA}
\bibfield{author}{\bibinfo{person}{F. {Xue}}, \bibinfo{person}{A. {Ming}},
  \bibinfo{person}{M. {Zhou}}, {and} \bibinfo{person}{Y. {Zhou}}.}
  \bibinfo{year}{2019}\natexlab{}.
\newblock \showarticletitle{A Novel Multi-layer Framework for Tiny Obstacle
  Discovery}. In \bibinfo{booktitle}{\emph{IEEE International Conference on
  Robotics and Automation (ICRA)}}.
\newblock


\bibitem[{Xue} et~al\mbox{.}(2020)]%
        {tip}
\bibfield{author}{\bibinfo{person}{F. {Xue}}, \bibinfo{person}{A. {Ming}},
  {and} \bibinfo{person}{Y. {Zhou}}.} \bibinfo{year}{2020}\natexlab{}.
\newblock \showarticletitle{Tiny Obstacle Discovery by Occlusion-aware
  Multilayer Regression}.
\newblock \bibinfo{journal}{\emph{IEEE Transactions on Image Processing (TIP)}}
   \bibinfo{volume}{29} (\bibinfo{year}{2020}), \bibinfo{pages}{9373--9386}.
\newblock


\bibitem[Yang et~al\mbox{.}(2024)]%
        {depthanything}
\bibfield{author}{\bibinfo{person}{Lihe Yang}, \bibinfo{person}{Bingyi Kang},
  \bibinfo{person}{Zilong Huang}, \bibinfo{person}{Xiaogang Xu},
  \bibinfo{person}{Jiashi Feng}, {and} \bibinfo{person}{Hengshuang Zhao}.}
  \bibinfo{year}{2024}\natexlab{}.
\newblock \showarticletitle{Depth Anything: Unleashing the Power of Large-Scale
  Unlabeled Data}. In \bibinfo{booktitle}{\emph{IEEE Conference on Computer
  Vision and Pattern Recognition (CVPR)}}.
\newblock


\bibitem[Yin et~al\mbox{.}(2023)]%
        {Metric3d2023}
\bibfield{author}{\bibinfo{person}{Wei Yin}, \bibinfo{person}{Chi Zhang},
  \bibinfo{person}{Hao Chen}, \bibinfo{person}{Zhipeng Cai},
  \bibinfo{person}{Gang Yu}, \bibinfo{person}{Kaixuan Wang},
  \bibinfo{person}{Xiaozhi Chen}, {and} \bibinfo{person}{Chunhua Shen}.}
  \bibinfo{year}{2023}\natexlab{}.
\newblock \showarticletitle{Metric3D: Towards Zero-shot Metric 3D Prediction
  from A Single Image}. In \bibinfo{booktitle}{\emph{IEEE International
  Conference on Computer Vision (ICCV)}}.
\newblock


\bibitem[Yin et~al\mbox{.}(2022)]%
        {BoostingDepth2022}
\bibfield{author}{\bibinfo{person}{Wei Yin}, \bibinfo{person}{Jianming Zhang},
  \bibinfo{person}{Oliver Wang}, \bibinfo{person}{Simon Niklaus},
  \bibinfo{person}{Simon Chen}, \bibinfo{person}{Yifan Liu}, {and}
  \bibinfo{person}{Chunhua Shen}.} \bibinfo{year}{2022}\natexlab{}.
\newblock \showarticletitle{Towards accurate reconstruction of 3d scene shape
  from a single monocular image}.
\newblock \bibinfo{journal}{\emph{IEEE Transactions on Pattern Analysis and
  Machine Intelligence}} (\bibinfo{year}{2022}).
\newblock


\bibitem[Yu et~al\mbox{.}(2023)]%
        {10.1145/3581783.3612232}
\bibfield{author}{\bibinfo{person}{Chaohui Yu}, \bibinfo{person}{Qiang Zhou},
  \bibinfo{person}{Jingliang Li}, \bibinfo{person}{Zhe Zhang},
  \bibinfo{person}{Zhibin Wang}, {and} \bibinfo{person}{Fan Wang}.}
  \bibinfo{year}{2023}\natexlab{}.
\newblock \showarticletitle{Points-to-3D: Bridging the Gap between Sparse
  Points and Shape-Controllable Text-to-3D Generation}. In
  \bibinfo{booktitle}{\emph{Proceedings of the 31st ACM International
  Conference on Multimedia}}.
\newblock


\bibitem[Yuan et~al\mbox{.}(2022)]%
        {NeWCRFs2022}
\bibfield{author}{\bibinfo{person}{Weihao Yuan}, \bibinfo{person}{Xiaodong Gu},
  \bibinfo{person}{Zuozhuo Dai}, \bibinfo{person}{Siyu Zhu}, {and}
  \bibinfo{person}{Ping Tan}.} \bibinfo{year}{2022}\natexlab{}.
\newblock \showarticletitle{New crfs: Neural window fully-connected crfs for
  monocular depth estimation}.
\newblock \bibinfo{journal}{\emph{arXiv preprint arXiv:2203.01502}}
  (\bibinfo{year}{2022}).
\newblock


\bibitem[Zhan et~al\mbox{.}(2023)]%
        {10230895}
\bibfield{author}{\bibinfo{person}{Fangneng Zhan}, \bibinfo{person}{Yingchen
  Yu}, \bibinfo{person}{Rongliang Wu}, \bibinfo{person}{Jiahui Zhang},
  \bibinfo{person}{Shijian Lu}, \bibinfo{person}{Lingjie Liu},
  \bibinfo{person}{Adam Kortylewski}, \bibinfo{person}{Christian Theobalt},
  {and} \bibinfo{person}{Eric Xing}.} \bibinfo{year}{2023}\natexlab{}.
\newblock \showarticletitle{Multimodal Image Synthesis and Editing: The
  Generative AI Era}.
\newblock \bibinfo{journal}{\emph{IEEE Transactions on Pattern Analysis and
  Machine Intelligence}} \bibinfo{volume}{45}, \bibinfo{number}{12}
  (\bibinfo{year}{2023}), \bibinfo{pages}{15098--15119}.
\newblock


\bibitem[Zhang et~al\mbox{.}(2021)]%
        {10.1145/3474085.3475386}
\bibfield{author}{\bibinfo{person}{Jiehua Zhang}, \bibinfo{person}{Liang Li},
  \bibinfo{person}{Chenggang Yan}, \bibinfo{person}{Yaoqi Sun},
  \bibinfo{person}{Tao Shen}, \bibinfo{person}{Jiyong Zhang}, {and}
  \bibinfo{person}{Zhan Wang}.} \bibinfo{year}{2021}\natexlab{}.
\newblock \showarticletitle{Heuristic Depth Estimation with Progressive Depth
  Reconstruction and Confidence-Aware Loss}. In
  \bibinfo{booktitle}{\emph{Proceedings of the 29th ACM International
  Conference on Multimedia}}.
\newblock


\bibitem[Zhang et~al\mbox{.}(2023)]%
        {zhang2023adding}
\bibfield{author}{\bibinfo{person}{Lvmin Zhang}, \bibinfo{person}{Anyi Rao},
  {and} \bibinfo{person}{Maneesh Agrawala}.} \bibinfo{year}{2023}\natexlab{}.
\newblock \bibinfo{title}{Adding Conditional Control to Text-to-Image Diffusion
  Models}.
\newblock
\newblock


\bibitem[Zhang et~al\mbox{.}(2022)]%
        {10.1145/3503161.3549201}
\bibfield{author}{\bibinfo{person}{Renrui Zhang}, \bibinfo{person}{Ziyao Zeng},
  \bibinfo{person}{Ziyu Guo}, {and} \bibinfo{person}{Yafeng Li}.}
  \bibinfo{year}{2022}\natexlab{}.
\newblock \showarticletitle{Can Language Understand Depth?}. In
  \bibinfo{booktitle}{\emph{Proceedings of the 30th ACM International
  Conference on Multimedia}}.
\newblock


\bibitem[Zhenda et~al\mbox{.}(2023)]%
        {mim}
\bibfield{author}{\bibinfo{person}{Xie Zhenda}, \bibinfo{person}{Geng Zigang},
  \bibinfo{person}{Hu Jingcheng}, \bibinfo{person}{Zhang Zheng},
  \bibinfo{person}{Hu Han}, {and} \bibinfo{person}{Cao Yue}.}
  \bibinfo{year}{2023}\natexlab{}.
\newblock \showarticletitle{Revealing the Dark Secrets of Masked Image
  Modeling}. In \bibinfo{booktitle}{\emph{IEEE Conference on Computer Vision
  and Pattern Recognition (CVPR)}}.
\newblock


\bibitem[Zheng et~al\mbox{.}(2018)]%
        {10.1145/3240508.3240628}
\bibfield{author}{\bibinfo{person}{Kecheng Zheng}, \bibinfo{person}{Zheng-Jun
  Zha}, \bibinfo{person}{Yang Cao}, \bibinfo{person}{Xuejin Chen}, {and}
  \bibinfo{person}{Feng Wu}.} \bibinfo{year}{2018}\natexlab{}.
\newblock \showarticletitle{LA-Net: Layout-Aware Dense Network for Monocular
  Depth Estimation}. In \bibinfo{booktitle}{\emph{Proceedings of the 26th ACM
  International Conference on Multimedia}}.
\newblock


\bibitem[Zhou et~al\mbox{.}(2023)]%
        {zhou2023occlusion}
\bibfield{author}{\bibinfo{person}{Yu Zhou}, \bibinfo{person}{Rui Lu},
  \bibinfo{person}{Feng Xue}, {and} \bibinfo{person}{Yuzhe Gao}.}
  \bibinfo{year}{2023}\natexlab{}.
\newblock \showarticletitle{Occlusion relationship reasoning with a feature
  separation and interaction network}.
\newblock \bibinfo{journal}{\emph{Visual Intelligence}} \bibinfo{volume}{1},
  \bibinfo{number}{1} (\bibinfo{year}{2023}), \bibinfo{pages}{23}.
\newblock


\bibitem[Zhou and Dong(2022)]%
        {10.1145/3503161.3548381}
\bibfield{author}{\bibinfo{person}{Zhengming Zhou} {and}
  \bibinfo{person}{Qiulei Dong}.} \bibinfo{year}{2022}\natexlab{}.
\newblock \showarticletitle{Learning Occlusion-aware Coarse-to-Fine Depth Map
  for Self-supervised Monocular Depth Estimation}. In
  \bibinfo{booktitle}{\emph{Proceedings of the 30th ACM International
  Conference on Multimedia}}.
\newblock


\end{thebibliography}

\balance

\end{document}